\documentclass{article}
\usepackage{preprint,times}
\usepackage[T1]{fontenc}
\usepackage{microtype}

\usepackage{amsmath,amsfonts,bm}
\usepackage{xcolor}

\usepackage{url}
\usepackage{algorithm}
\usepackage{algpseudocode}
\usepackage{amssymb}
\usepackage{mathtools}
\usepackage{tabularx}
\usepackage{multirow}
\usepackage{colortbl}
\usepackage{arydshln}
\definecolor{softcyan}{HTML}{CDEFF1}
\usepackage{caption}
\usepackage{booktabs}
\usepackage{graphicx}
\usepackage{titletoc}
\usepackage{listings}
\usepackage{tcolorbox}
\tcbuselibrary{listings,breakable,skins}
\newcommand{\method}{\textsc{CoMixKV}}
\definecolor{softgray}{HTML}{F2F2F2}
\definecolor{promptTealBlue}{cmyk}{0.86,0,0.34,0.02}
\definecolor{promptPeriwinkle}{cmyk}{0.57,0.55,0,0}
\definecolor{promptForestGreen}{cmyk}{0.91,0,0.88,0.12}
\definecolor{promptshadow}{RGB}{200,200,200}
\newcommand{\fullkvcell}[1]{\cellcolor{softgray}#1}
\tcbset{
  promptboxbase/.style={
    enhanced,
    breakable,
    sharp corners,
    colback=white,
    coltitle=black,
    boxrule=3pt,
    boxsep=0.5pt,
    shadow={3pt}{-3pt}{0pt}{opacity=1,promptshadow},
    fonttitle=\bfseries,
    listing only,
    listing engine=listings,
    listing options={
      basicstyle=\ttfamily\footnotesize,
      breaklines=true,
      breakatwhitespace=true,
      columns=fullflexible,
      keepspaces=true,
      showstringspaces=false,
      aboveskip=0pt,
      belowskip=0pt
    }
  }
}
\newtcblisting{prompttealbox}[1]{promptboxbase,colframe=promptTealBlue,colbacktitle=promptTealBlue!15!white,title={#1}}
\newtcblisting{promptperiwinklebox}[1]{promptboxbase,colframe=promptPeriwinkle,colbacktitle=promptPeriwinkle!15!white,title={#1}}
\newtcblisting{promptgreenbox}[1]{promptboxbase,colframe=promptForestGreen,colbacktitle=promptForestGreen!15!white,title={#1}}

\usepackage{hyperref}
\hypersetup{
    colorlinks=true,
    linkcolor=red,
    citecolor=cyan,
    urlcolor=blue,
    pdftitle={Coordinated Mixed-Precision KV Cache Quantization for Efficient LLM-based Multi-agent Systems},
    pdfauthor={Weizhong Huang, Jinchao Zhang, Xiawu Zheng}
}

\title{Coordinated Mixed-Precision KV Cache Quantization for Efficient LLM-based Multi-agent Systems}

\author{
  Weizhong Huang\textsuperscript{1,\textdagger}\quad
  Jinchao Zhang\textsuperscript{2,*}\quad
  Xiawu Zheng\textsuperscript{1,*}\\[0.5em]
  {\normalsize\textsuperscript{1}Xiamen University\quad
  \textsuperscript{2}WeChat AI, Tencent Inc., China}\\[0.3em]
  {\small\textsuperscript{*}Corresponding authors}
}
\date{}

\begin{document}

\maketitle
\begingroup
\renewcommand{\thefootnote}{\fnsymbol{footnote}}
\footnotetext[2]{This work was done when Weizhong Huang was an intern at WeChat AI.}
\endgroup

\begin{abstract}
LLM-based multi-agent systems solve complex tasks through problem decomposition, specialized tool use, and iterative verification. However, collaboration produces multiple growing key--value (KV) caches that can coexist in memory, making cache storage a bottleneck. KV-cache quantization reduces this footprint, but independently applying single-model policies ignores differences in agents' precision needs. This can over-compress sensitive computations while spending precision on more tolerant caches, potentially degrading collective reasoning. In this paper, we introduce \textbf{Co}ordinated \textbf{Mix}ed-Precision \textbf{KV} Cache Quantization (\method{}), a novel method that integrates online sensitivity estimation with joint bit-width allocation under a shared memory budget. To our knowledge, \method{} is the first mixed-precision KV-cache quantization framework to jointly allocate precision across simultaneously resident agent caches under such a budget. An online estimator derives each agent's sensitivity from a local attention-error bound using recent queries, keys, and values, without calibration data, trial quantization, or additional model passes. A shared controller combines these scores with remaining quantizable capacity and byte-level storage costs to minimize total estimated error. Allocations are updated as agents start or finish and fresh observations become available. Selected widths apply only to newly quantized blocks; recent states retain native precision and previously quantized blocks remain unchanged. Across four benchmarks and three LLM backbones at average 2-bit, 3-bit, and 4-bit KV-cache settings, \method{} achieves strong accuracy compared with the evaluated quantization baselines. At the average 3-bit setting on Gemma-4-31B-it, \method{} achieves 68.35\% average accuracy, outperforming the strongest quantized baseline by 0.59\%. On Qwen3-32B, \method{} further reduces peak KV-cache memory by 32.08\% and achieves a 1.38$\times$ inference speedup over FullKV.
\end{abstract}

\section{Introduction}

Large language model (LLM)-based multi-agent systems solve complex tasks through collaboration among specialized agents~\citep{guo2024large,li2024survey,chen2024survey}. Agents take complementary roles, use tools, and exchange intermediate results~\citep{yang2024llm,he2025llm,wu2024autogen,metagpt,fourney2024magentic}. This collaboration, however, increases memory consumption because each agent maintains a local history whose attention states are stored in a key--value (KV) cache. Distinct histories form separate cache branches, and overlapping agent lifetimes can make multiple caches resident simultaneously, substantially increasing aggregate KV storage~\citep{pan2025kvflow,wang2025kvcache,gao2024cost,jin2025compute}.

KV-cache quantization reduces this footprint by storing keys and values at lower precision~\citep{kvquant,hariri2026quantize,zhang2025more}. Representative methods include \textbf{KIVI}~\citep{kivi}, \textbf{PolarQuant}~\citep{wu2026polarquant}, \textbf{TurboQuant}~\citep{zandieh2026turboquant}, \textbf{PM-KVQ}~\citep{liu2026pm}, and \textbf{PatternKV}~\citep{zhang2026patternkv}. These methods improve quantization representations or precision allocation within individual caches~\citep{kivi,zipcache,kvquant,aq_mixkvq,zuo2026ratequant,shen2026triaxialkv}, but do not coordinate precision across simultaneously resident agent caches.

Applying such policies independently leaves two challenges. First, different agents and layers process different queries and histories, so the same precision can produce substantially different attention errors. Uniform or fixed allocation may therefore waste precision on tolerant computations while under-protecting sensitive ones. Second, cache residency changes as agents start, pause, and finish. Since previously quantized blocks retain their precision, later decisions affect only future blocks and must account for both remaining quantizable capacity and already committed storage. Effective compression therefore requires coordinated precision allocation across changing agent--layer computations under a shared memory budget.

We introduce \textbf{Co}ordinated \textbf{Mix}ed-Precision \textbf{KV} Cache Quantization (\method{}), which combines online sensitivity estimation with agent--layer bit-width allocation across private resident caches. To our knowledge, \method{} is the first mixed-precision KV-cache quantization framework to coordinate precision across simultaneously resident agent caches under a shared memory budget. Agents exchange text while retaining private KV states, and the controller coordinates precision rather than transferring cache tensors.

\method{} estimates agent--layer quantization sensitivity from recent queries, keys, and values produced during ordinary inference. The score is derived from a local attention-error bound and requires no calibration data, trial quantization, or additional model passes. Based on these scores, a shared controller assigns lower-, target-, and higher-bit widths across layers while respecting the byte budget. More sensitive layers receive higher precision, while balanced assignments preserve the target mean packed-code width. Allocations are refreshed as agent residency and observations change, and new assignments apply only to newly quantized blocks; existing blocks retain their stored precision.

We evaluate \method{} on four benchmarks spanning knowledge search, agentic reasoning, mathematical reasoning, and scientific reasoning, using three LLM backbones and average 2-, 3-, and 4-bit KV-cache settings. Against three KV-cache quantization baselines, \method{} achieves strong accuracy across the evaluated settings. At the average 3-bit setting on Gemma-4-31B-it, \method{} achieves 68.35\% average accuracy, outperforming the strongest quantized baseline by 0.59\%. Moreover, on Qwen3-32B, \method{} reduces peak KV-cache memory by 32.08\% and achieves a 1.38$\times$ inference speedup over FullKV, demonstrating improvements in both task accuracy and system efficiency.

\section{Related Work}

\paragraph{LLM-based multi-agent systems.}
LLM agents interleave reasoning, tool use, and environment feedback across turns~\citep{react,masterman2024landscape,zhang2025tool}. Multi-agent systems extend this process through role-based collaboration~\citep{metagpt,wu2024autogen,li2023camel,qian2024chatdev} and verification~\citep{zhang2026verified}. Their growing local histories and overlapping cache lifetimes increase aggregate KV-cache costs. Recent methods address this burden through cross-context reuse~\citep{ye2026kvcomm}, workflow-guided retention and prefetching~\citep{pan2025kvflow}, coordinated scheduling and offloading~\citep{aq_tokencake}, or shared compressed pools~\citep{aq_polykv}. \method{} instead coordinates precision across private resident caches under a shared memory budget. Agents exchange textual reports rather than KV states, while their starts and returns change which caches compete for memory.

\paragraph{KV cache quantization.}
KV cache compression~\citep{h2o,scissorhands,quest,adakv,ge2024model} is an effective method for reducing long-context inference costs~\citep{cam,minicache,pagedeviction,take}; specifically, KV cache quantization~\citep{aq_qkvcomm,aq_qkvshare} reduces memory use without removing tokens through outlier handling~\citep{kvquant}, adaptive precision~\citep{zipcache,qaq,aq_mixkvq,zuo2026ratequant}, and alternative representations~\citep{pqcache,gear}. KIVI~\citep{kivi} quantizes keys channel-wise and values token-wise, retaining native-precision tails. PolarQuant~\citep{wu2026polarquant} uses polar quantization for RoPE-coupled keys; TurboQuant~\citep{zandieh2026turboquant} combines random rotations, scalar quantization, and residual correction. PM-KVQ~\citep{liu2026pm} reduces precision with sequence growth using calibrated layer sensitivity; PatternKV~\citep{zhang2026patternkv} quantizes residuals from online-mined patterns. TriAxialKV~\citep{shen2026triaxialkv} calibrates precision by recency, modality, and semantic role. \method{} uses online Q/K/V sensitivity and exact per-invocation packed-code balance within a shared cache budget.

\section{Preliminaries}
\label{sec:preliminaries}

We describe multi-agent inference as a spatial-temporal graph and use this view to formulate KV cache quantization over the resulting execution structure.

\subsection{Multi-Agent Inference as a Spatial-Temporal Graph}
\label{sec:graph_formulation}

\paragraph{Execution graph.}
Given a query $\boldsymbol{q}$, we represent multi-agent execution as a directed spatial-temporal graph~\citep{agentprune,gptswarm,dylan}:
\begin{equation}
\mathcal{G}_{\boldsymbol{q}}=(\mathcal{V}_{\boldsymbol{q}},\mathcal{E}_{\boldsymbol{q}}),
\qquad
\mathcal{E}_{\boldsymbol{q}}=\mathcal{E}_{\boldsymbol{q}}^{\mathrm{S}}\cup\mathcal{E}_{\boldsymbol{q}}^{\mathrm{T}}.
\label{eq:agent_graph}
\end{equation}
The node set $\mathcal{V}_{\boldsymbol{q}}$ contains execution states $v_i^{(t)}$, each representing agent $i$ at step $t$:
\begin{equation}
v_i^{(t)}=\left\{\texttt{Base}_i,\texttt{Role}_i,\texttt{State}_i^{(t)},\texttt{Tools}_i\right\},
\label{eq:agent_node}
\end{equation}
where $\texttt{Base}_i=\mathcal{M}$ is the language model, and $\texttt{Role}_i$, $\texttt{State}_i^{(t)}$, and $\texttt{Tools}_i$ specify its responsibility, local interaction history, and available tools, respectively~\citep{wu2024autogen}. Spatial edges in $\mathcal{E}_{\boldsymbol{q}}^{\mathrm{S}}$ carry task assignments, worker reports, or verification feedback between agents~\citep{wu2024autogen,agentscope}, while temporal edges in $\mathcal{E}_{\boldsymbol{q}}^{\mathrm{T}}$ carry information from earlier to later execution states. All edges follow runtime order, forming a directed acyclic graph (DAG)~\citep{bondy1976graph}.

\begin{figure}[t]
    \centering
    \includegraphics[width=\linewidth]{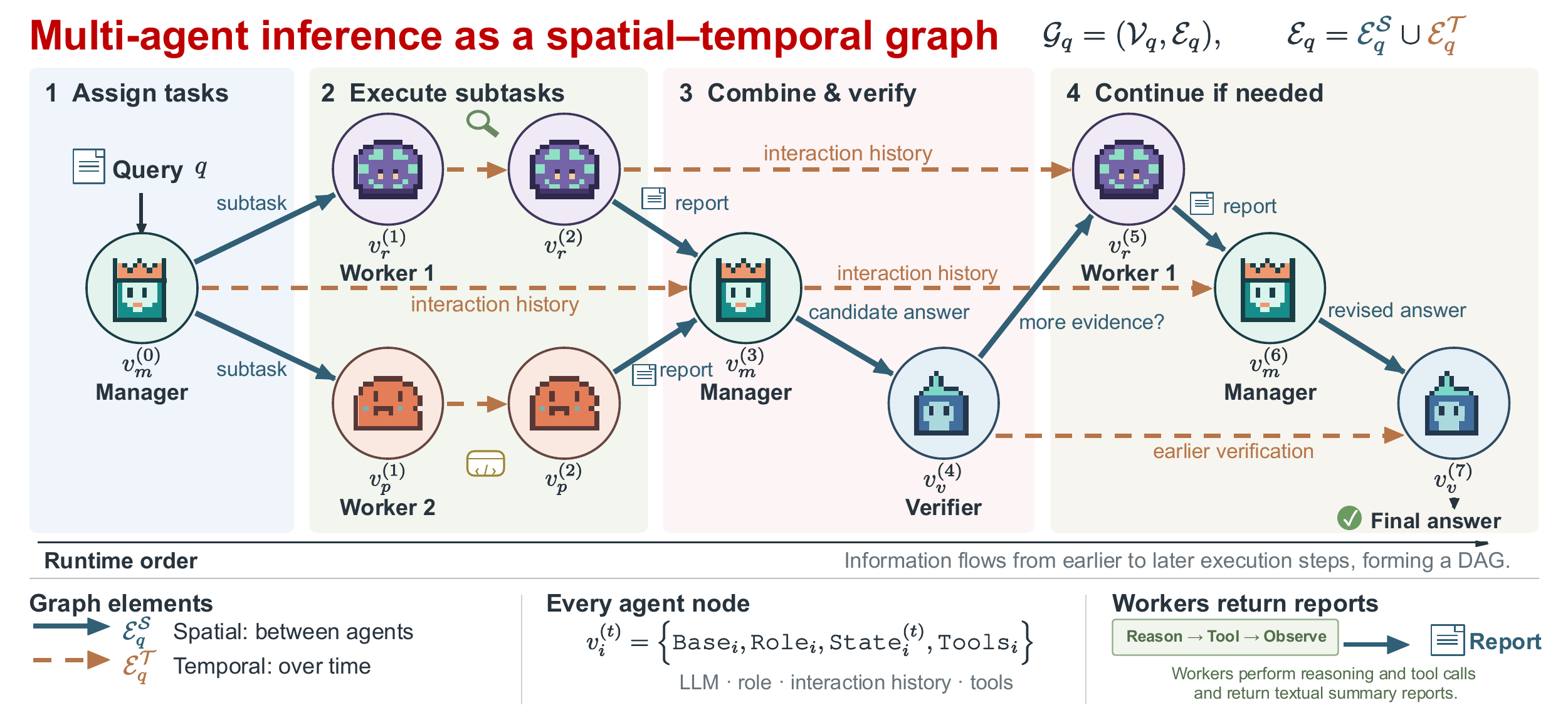}
    \caption{\textbf{Multi-agent inference as a spatial-temporal graph.} Solid blue arrows denote spatial communication ($\mathcal{E}_{\boldsymbol{q}}^{\mathrm{S}}$), while dashed orange arrows denote temporal information flow ($\mathcal{E}_{\boldsymbol{q}}^{\mathrm{T}}$). Node subscripts $m$, $r$, $p$, and $v$ denote the manager, Worker 1, Worker 2, and verifier, respectively.}
    \label{fig:multi_agent_graph}
\end{figure}

\paragraph{Multi-agent communication.}
Each agent follows a tool-calling ReAct process~\citep{react}, generating its output from the query, role, local history, and incoming temporal and spatial information:
\begin{equation}
\boldsymbol{y}_i^{(t)}\sim\mathcal{P}\left(\cdot\mid\boldsymbol{q},\texttt{Role}_i,\texttt{State}_i^{(t)},\boldsymbol{x}_i^{\mathrm{T},(t)},\boldsymbol{x}_i^{\mathrm{S},(t)}\right),
\label{eq:agent_communication}
\end{equation}
where $\mathcal{P}$ is the model's conditional output distribution, and $\boldsymbol{x}_i^{\mathrm{T},(t)}$ and $\boldsymbol{x}_i^{\mathrm{S},(t)}$ denote earlier-state information and messages from other agents, respectively. The output $\boldsymbol{y}_i^{(t)}$ may include reasoning, tool calls, task assignments, reports, or a final answer.

In our hierarchical workflow, as shown in Figure~\ref{fig:multi_agent_graph}, the manager delegates subtasks to specialized workers~\citep{fourney2024magentic}, which reason and use tools while keeping detailed histories local and returning only textual reports. The manager combines these reports into a candidate answer for the verifier. When more evidence is needed, the runtime starts another worker invocation and supplies its report, together with the original candidate answer, to a fresh verifier invocation. These exchanges form related execution branches rather than a single linear history.

\subsection{Graph-Conditioned KV-Cache Quantization}
\label{sec:graph_conditioned_quantization}

\paragraph{Graph-induced cache branches.}
Each invocation $a$ of agent $i$ maintains a local history $\boldsymbol{h}_a$, formed by the inputs in Eq.~(\ref{eq:agent_communication}) and its generated tokens. The corresponding keys and values are stored in cache~\citep{vllm,kivi,kvquant}:
\begin{equation}
\mathcal{KV}_a=\bigl\{(\boldsymbol{K}_a^{(\ell)},\boldsymbol{V}_a^{(\ell)})\bigr\}_{\ell\in\mathcal{L}},
\qquad \boldsymbol{K}_a^{(\ell)},\boldsymbol{V}_a^{(\ell)}\in\mathbb R^{N_a\times D_\ell}.
\label{eq:aq_agent_cache}
\end{equation}
Here, $N_a$ counts cached tokens, $\mathcal{L}$ contains the attention layers to be quantized, and $D_\ell$ is the total feature dimension across KV heads in layer $\ell$.

\begin{figure}[t]
    \centering
    \includegraphics[width=\linewidth]{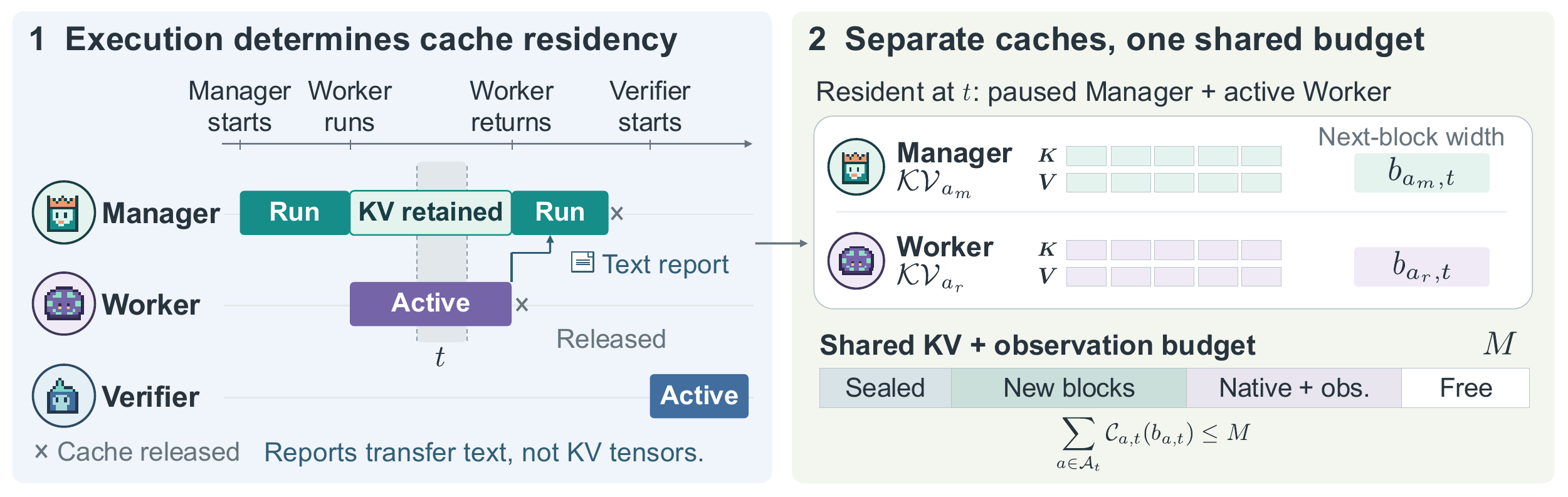}
    \caption{\textbf{Graph-conditioned cache residency and shared-budget quantization.} Left: cache lifetimes across manager, worker, and verifier runs. Right: agent--layer width choices for resident private KV caches under a shared byte envelope.}
    \label{fig:graph_conditioned_kv_cache}
\end{figure}

Temporal edges within an invocation extend its local history by retaining existing KV states and appending new entries~\citep{yu2025stateful,gao2024cost}, whereas spatial edges transfer messages that are incorporated into the receiver's local history and encoded into its own KV states. As shown in Figure~\ref{fig:graph_conditioned_kv_cache}, the manager retains its cache while a nested worker runs. The worker's cache is released on return, but its textual report remains in the manager's history. The manager's cache is released before the separate verifier run begins. Thus, execution order determines which cache branches coexist and share the memory budget.

\paragraph{Problem formulation.}
\label{sec:problem_formulation}

Eq.~(\ref{eq:aq_agent_cache}) defines $\mathcal{KV}_a$ as the keys and values stored across attention layers for one agent invocation $a$. Each invocation maintains the KV states computed from its own local history. Several such caches can occupy memory at the same time, even when the agents are not generating tokens simultaneously. For example, the manager keeps its cache while waiting for a worker's report, and the worker maintains its own cache during this period. Both caches therefore need to be considered when allocating the available memory.

We jointly choose the bit widths for these caches and update the choices during execution. Each such decision is called an \emph{allocation round}. A new round can occur when an invocation starts or finishes, or when new observations become available for estimating quantization sensitivity. At allocation round $r$, let $\mathcal{A}_r$ contain all invocations whose caches are still stored in memory, including invocations waiting for other agents to return. Collecting their individual caches $\mathcal{KV}_a$ gives
\begin{equation}
\mathcal{KV}^{(r)}=\bigl\{\mathcal{KV}_a:a\in\mathcal{A}_r\bigr\}.
\label{eq:aq_resident_caches}
\end{equation}
Each element of $\mathcal{KV}^{(r)}$ is therefore the cache of one invocation, containing its keys and values across attention layers. These caches remain separate, but we select the bit widths for their newly quantized blocks jointly, accounting for their quantization sensitivities and storage requirements under a shared memory budget. The controller and budget are scoped to one question's agent team; unrelated questions use separate controller states.

For each invocation, we keep the KV states of the most recent $R$ tokens at the model's original precision. In each layer, older states are grouped into blocks of $P$ consecutive tokens for quantization. We set a target packed-code mean $b_{\mathrm{avg}}\in\{2,3,4\}$ and allow three choices: $\mathcal{B}=\{b_{\mathrm{avg}}-1,b_{\mathrm{avg}},b_{\mathrm{avg}}+1\}$. Thus, the audited 2-, 3-, and 4-bit runs use candidates $\{1,2,3\}$, $\{2,3,4\}$, and $\{3,4,5\}$, respectively. The target and candidate set remain fixed throughout an experiment; the target is the mean packed-code width, while individual sealed blocks may use any candidate.

At allocation round $r$, we choose $b_{a,\ell,r}\in\mathcal{B}$ for each invocation $a$ and KV layer $\ell$. Until the next round, this width applies to newly quantized key and value blocks in that layer. Already quantized blocks, called \emph{sealed} blocks, keep their existing widths. Let $\mathcal{B}_{a,j}^{(\ell)}=(\boldsymbol{K}_{a,j}^{(\ell)},\boldsymbol{V}_{a,j}^{(\ell)})$ contain the keys and values of the $j$-th block from invocation $a$ in layer $\ell$. When first quantized using the width selected at round $r$, the block is stored as
\begin{equation}
\overline{\mathcal{B}}_{a,j}^{(\ell)}=\mathcal{Q}_{b_{a,\ell,r}}\!\left(\mathcal{B}_{a,j}^{(\ell)}\right).
\label{eq:aq_agent_block_quantization}
\end{equation}
Here, $\mathcal{Q}_b$ quantizes keys and values directly at the selected $b$-bit resolution and stores the codes and parameters needed to reconstruct them for attention. The packed implementation stores a fixed base payload plus refinement bitplanes for widths above that base; physical refinement pages are committed on demand. In the audited average 2-bit runs the base is one bit, allowing 1-bit blocks to offset the bytes of 3-bit blocks. The average 3- and 4-bit runs use 2- and 3-bit bases, respectively. Following KIVI~\citep{kivi}, keys are grouped across tokens within each feature, while values are grouped across features within each token. Let $T_a$ be the cache capacity in tokens and $z_{a,\ell,r}$ the number of already quantized tokens in layer $\ell$. The number of additional quantizable blocks in that layer is
\begin{equation}
n_{a,\ell,r}=\max\left(0,\left\lfloor\frac{T_a-R}{P}\right\rfloor-\frac{z_{a,\ell,r}}{P}\right).
\label{eq:aq_remaining_blocks}
\end{equation}
This counts remaining quantizable capacity, not predicted generation length. Let $s_{a,\ell,r}\ge0$ denote the sensitivity of layer $\ell$ computed from a small window of recent queries, keys, and values. The allocation surrogate for a candidate assignment $\boldsymbol{b}=(b_{a,\ell})$ is
\begin{equation}
\widehat{\mathcal{E}}_r(\boldsymbol{b})=\sum_{a\in\mathcal{A}_r}\sum_{\ell\in\mathcal{L}}\frac{s_{a,\ell,r}}{(2^{b_{a,\ell}}-1)^2}.
\label{eq:aq_estimated_error}
\end{equation}
The estimate treats recent sensitivity as representative of future blocks; it does not measure their actual future error. The audited allocation chooses each agent--layer vector under exact per-invocation balance and full three-width coverage:
\begin{equation}
\boldsymbol{b}_r\in\operatorname*{arg\,min}_{\boldsymbol{b}\in\mathcal{B}^{|\mathcal{A}_r|L}}\widehat{\mathcal{E}}_r(\boldsymbol{b})\quad\text{s.t.}\quad \sum_{\ell\in\mathcal{L}}b_{a,\ell}=Lb_{\mathrm{avg}},\ \{b_{a,\ell}:\ell\in\mathcal{L}\}=\mathcal{B}\ (\forall a),\ \sum_{a\in\mathcal{A}_r}\mathcal{C}_{a,r}(\boldsymbol{b}_a)\le M_r^{\mathrm{eff}}.
\label{eq:aq_problem}
\end{equation}
The implementation strengthens the mean constraint to equal counts of $b_{\mathrm{avg}}-1$ and $b_{\mathrm{avg}}+1$ layers for each invocation, leaving at least one layer at $b_{\mathrm{avg}}$. Equal per-layer KV geometry makes this balance exact for each complete group written across layers. Sealed blocks contribute to memory cost but not the prospective objective because their precision cannot change. The selection rule and physical audit are specified below.

\section{Methodology}
\label{sec:method}

At allocation round $r$, the formulation above requires a sensitivity score $s_{a,\ell,r}$ for each agent and layer, and a byte reservation $\mathcal{C}_{a,r}(\boldsymbol{b}_a)$ for each proposed layer vector. We obtain these quantities during ordinary inference, choose balanced layer vectors under the shared byte budget, and apply them to newly quantized cache blocks. Different agents may choose different layer vectors without a separate calibration dataset.

\begin{figure}[t]
    \centering
    \includegraphics[width=\linewidth]{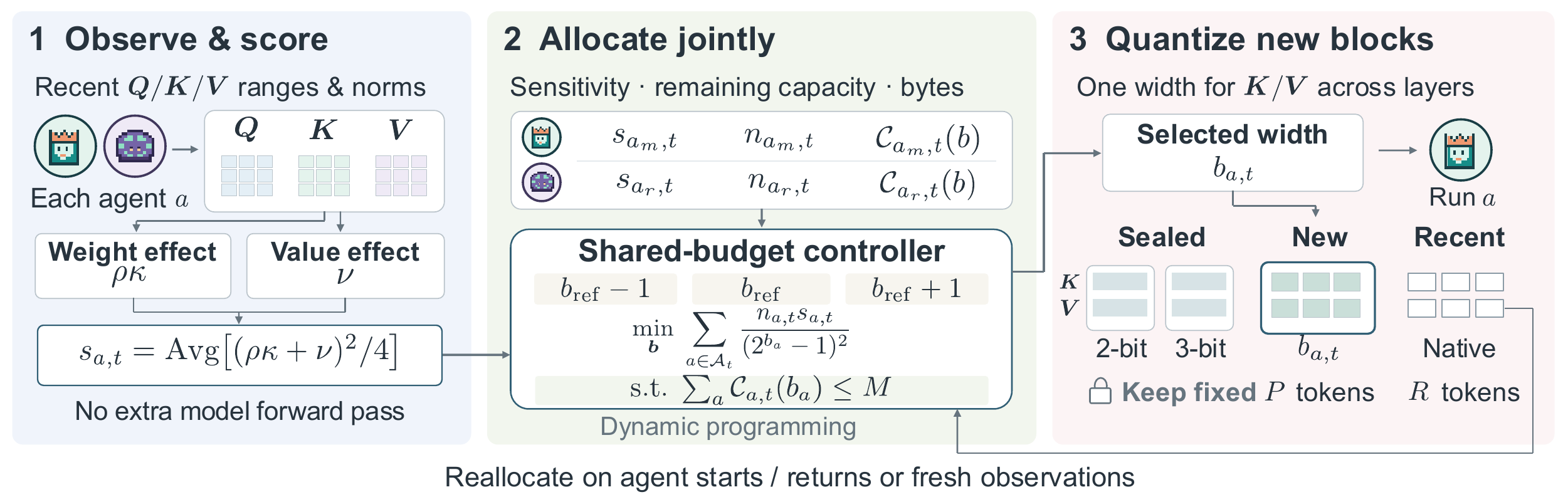}
    \caption{\textbf{Overview of \method{}.} \textbf{Observe and score:} recent $\boldsymbol{Q}/\boldsymbol{K}/\boldsymbol{V}$ observations provide online sensitivity scores by agent and layer. \textbf{Allocate:} the controller assigns paired lower- and higher-bit layers with the target packed-code mean for each invocation under the shared byte envelope. \textbf{Quantize new blocks:} selected widths apply only to newly quantized blocks; sealed blocks remain unchanged, and the most recent $R$ tokens retain native precision. Allocations are updated when the resident set changes or fresh observations become available.}
    \label{fig:comixkv_overview}
\end{figure}

\subsection{Compute Sensitivity from Recent Queries, Keys, and Values}

We compute $s_{a,\ell,r}$, the sensitivity score used in Eq.~(\ref{eq:aq_estimated_error}), from recent queries, keys, and values of invocation $a$ in layer $\ell$. Instead of using the entire history, we collect a small recent observation window. Recent queries provide a compact signal of the agent's current attention behavior, while restricting the computation to a short window keeps the online estimation inexpensive. Similar observation windows have been used in KV-cache compression to estimate token importance from recent attention patterns~\citep{snapkv,pyramidkv,adakv}; here the score estimates sensitivity to quantization.

For one attention head, let $\boldsymbol{q}\in\mathbb R^d$ be an attention query vector and let $\boldsymbol{K},\boldsymbol{V}\in\mathbb R^{w\times d}$ contain $w$ recent KV entries with $d$ features each. The attention output over this window is
\begin{equation}
\boldsymbol{o}=\operatorname{Softmax}\left(\boldsymbol{K}\boldsymbol{q}/\sqrt{d}\right)^{\top}\boldsymbol{V}.
\label{eq:aq_attention}
\end{equation}
Softmax converts the query--key scores into weights that sum to one, and the resulting attention output is $\boldsymbol{o}\in\mathbb R^d$. Following KIVI~\citep{kivi}, we group keys across tokens within each feature and values across features within each token. A group's range is its maximum minus its minimum; a wider range allows larger rounding errors at the same bit width. We compute three scalars directly from $\boldsymbol{q}$, $\boldsymbol{K}$, and $\boldsymbol{V}$:
\begin{equation}
\resizebox{\linewidth}{!}{$\displaystyle
\kappa=\frac{1}{\sqrt{d}}\sum_j|q_j|\max_{\mathcal{G}}\bigl(\max_{i\in\mathcal{G}}[\boldsymbol{K}]_{ij}-\min_{i\in\mathcal{G}}[\boldsymbol{K}]_{ij}\bigr),
\nu=\max_i\sqrt{\sum_{\mathcal{G}}|\mathcal{G}|\bigl(\max_{j\in\mathcal{G}}[\boldsymbol{V}]_{ij}-\min_{j\in\mathcal{G}}[\boldsymbol{V}]_{ij}\bigr)^2},
\rho=\frac{1}{2}\sqrt{\sum_j\bigl(\max_i[\boldsymbol{V}]_{ij}-\min_i[\boldsymbol{V}]_{ij}\bigr)^2}.
$}
\label{eq:aq_kappa_nu}
\end{equation}
Here, $[\boldsymbol{K}]_{ij}$ and $[\boldsymbol{V}]_{ij}$ denote matrix entries, with $i$ indexing tokens and $j$ indexing features, and $q_j$ is the corresponding query entry. In $\kappa$, $\mathcal{G}$ ranges over token groups; in $\nu$, it ranges over feature groups, with $|\mathcal{G}|$ counting the features in each group. The scalar $\kappa$ combines query magnitudes and key ranges to capture how rounding can change the query--key scores. The scalar $\nu$ captures the possible effect of value rounding, while $\rho$ measures how much the value vectors differ across tokens.

A $b$-bit uniform quantizer divides each group's range into $2^b-1$ equal intervals. With exact group endpoints and nearest rounding, each entry's error is at most half an interval. Using $\kappa,\nu,\rho$ from Eq.~(\ref{eq:aq_kappa_nu}), we combine the effects of key and value rounding to obtain
\begin{equation}
\|\widetilde{\boldsymbol{o}}_b-\boldsymbol{o}\|_2\le\frac{\rho\kappa+\nu}{2(2^b-1)}.
\label{eq:aq_attention_bound}
\end{equation}
Here, $\widetilde{\boldsymbol{o}}_b\in\mathbb R^{1\times d}$ is the attention output computed from reconstructed $b$-bit keys and values, with the query and visible positions unchanged. The numerator depends on the observed queries, keys, and values, while the denominator shows how increasing precision reduces the bound.

Squaring this bound separates the coefficient $(\rho\kappa+\nu)^2/4$ from the bit-dependent factor $(2^b-1)^{-2}$. We average the coefficient to obtain the sensitivity score for invocation $a$ at allocation round $r$:
\begin{equation}
s_{a,\ell,r}=\operatorname{Average}_{h,u}\!\left[\frac{(\rho\kappa+\nu)^2}{4}\right].
\label{eq:aq_online_score}
\end{equation}
Here, $h$ and $u$ index query heads and positions in layer $\ell$. Within the average, $\kappa$ uses the corresponding query and key head, while $\rho$ and $\nu$ use the associated value head; their indices are omitted for brevity. Query heads sharing one KV head use the same keys and values. A larger $s_{a,\ell,r}$ indicates larger estimated attention error at the same bit width. Summing $s_{a,\ell,r}/(2^{b_{a,\ell}}-1)^2$ over layers and resident invocations gives Eq.~(\ref{eq:aq_estimated_error}). The bound itself applies only to the observed window; using it for future blocks is a planning surrogate. Appendix~\ref{app:aq_attention_to_objective} gives the derivation and assumptions.

\subsection{Allocate the Shared Memory Budget}

At allocation round $r$, the memory reserved for invocation $a$ when its remaining quantizable blocks use layer vector $\boldsymbol{b}_a$ is
\begin{equation}
\mathcal{C}_{a,r}(\boldsymbol{b}_a)=I_{a,r}+\sum_{\ell\in\mathcal{L}}n_{a,\ell,r}\mathcal{B}_{\mathrm{blk}}(b_{a,\ell})+U_{a,r}+O_a,
\label{eq:aq_reservation_main}
\end{equation}
where $I_{a,r}$ is the storage of already quantized blocks and $n_{a,\ell,r}$ is remaining block capacity in layer $\ell$. The per-layer block cost $\mathcal{B}_{\mathrm{blk}}(b)$ includes packed codes, scales, and offsets. The terms $U_{a,r}$ and $O_a$ cover unquantized KV states and sensitivity observations, respectively. Appendix~\ref{app:aq_storage_details} gives the byte calculation.

The audited mixed-width runs enforce a cumulative physical-byte envelope at every allocation round. Let
\begin{equation}
\mathcal{C}^{\mathrm{tar}}_{a,r}
=\sum_{\ell\in\mathcal{L}}\left(\frac{z_{a,\ell,r}}{P}+n_{a,\ell,r}\right)\mathcal{B}_{\mathrm{blk}}(b_{\mathrm{avg}})+U_{a,r}+O_a,
\qquad
M_r^{\mathrm{eff}}=\min\left(M,\sum_{a\in\mathcal{A}_r}\mathcal{C}^{\mathrm{tar}}_{a,r}\right).
\label{eq:aq_target_envelope}
\end{equation}
The target entitlement charges sealed and prospective blocks at the average width, whereas $\mathcal{C}_{a,r}(\boldsymbol{b}_a)$ charges sealed blocks at their immutable widths. Pairing one lower-bit and one higher-bit layer within the same invocation keeps prospective payload bytes at the average target for equal-sized layer blocks. This avoids depending on byte savings from another agent that might later leave. The controller records whether each proposed reservation fits $M_r^{\mathrm{eff}}$, and the final audit also checks actual sealed-code widths and physical storage.

For each invocation, we sort layers by their current scores. For a fixed number $k$ of paired exchanges, the $k$ least-sensitive layers use $b_{\mathrm{avg}}-1$, the $k$ most-sensitive use $b_{\mathrm{avg}}+1$, and the remaining layers use $b_{\mathrm{avg}}$. We evaluate all $1\le k\le\lfloor(L-1)/2\rfloor$ and select the minimum surrogate in Eq.~(\ref{eq:aq_estimated_error}). The first pair guarantees all three widths; more pairs are selected only when they reduce the surrogate. If two invocations would receive the same vector, a minimum-cost swap between two differently coded layers makes their vectors distinct without changing either invocation's width counts. Appendix~\ref{app:aq_allocation_details} gives the selection details.

\subsection{Apply and Update the Bit-Width Assignment}

The selected width $b_{a,\ell,r}$ applies to newly quantized $\boldsymbol{K}$ and $\boldsymbol{V}$ blocks in layer $\ell$. New KV entries are first appended at native precision. After each forward pass, older unquantized entries outside the recent $R$-token window are quantized in complete blocks of $P$ tokens using the current assignment. Incomplete blocks remain unquantized. Each quantized block retains its stored width and codes, so later assignments affect only newly quantized blocks.

The controller updates assignments when an invocation starts or finishes, or when fresh sensitivity observations become available. Observations are collected during prefill and approximately every $J$ newly generated tokens. Before an invocation has a valid observation for a layer in its current request, that layer uses the median valid score among resident invocations (or $1$ if none exists) as a temporary planning score; its first observation replaces this value. Completed invocations release their caches. If a changed request invalidates a cached suffix, we discard it and recompute the assignment before rebuilding that suffix. Between updates, the current assignment is reused. The audited reruns use no score smoothing or switch hysteresis. The runtime records physically sealed widths by agent and layer for post-run checks, including the exact mean packed-code width.

Algorithm~\ref{alg:aq_online_allocation} presents the end-to-end procedure, integrating cache initialization and reuse, sensitivity estimation, joint bit-width allocation, block quantization, and cache release.

\section{Experiments}\label{sec:experiments}
\subsection{Experimental Setup}

\paragraph{Baselines.}
We compare \method{} with \textbf{KIVI}~\citep{kivi}, \textbf{TurboQuant}~\citep{zandieh2026turboquant}, and \textbf{PM-KVQ}~\citep{liu2026pm}. All the methods are representative KV-cache quantization baselines. For fair comparison, we use the same model and decoding configuration for all methods and set the same target bit width for every quantization baseline.

\paragraph{Models.}
We evaluate three open-weight backbones with dense and hybrid-attention architectures: Qwen3-32B~\citep{yang2025qwen3}, Qwen3.8-27B~\citep{qwen38_27b}, and Gemma-4-31B-it~\citep{team2026gemma}. This selection lets us examine \method{} across distinct model families and attention architectures.

\begin{table}[t!]
\centering
\captionsetup{skip=2pt}
\fontsize{9.0}{9.3}\selectfont
\setlength{\tabcolsep}{1.8pt}
\renewcommand{\arraystretch}{1.15}

\caption{Accuracy (\%) across four benchmarks and two model backbones at an average 3-bit KV-cache bit-width budget.}
\label{tab:main_results}

\resizebox{\textwidth}{!}{%
\begin{tabular}{@{}llccccc@{}}
\toprule
\multirow{2}{*}{\shortstack[l]{\textbf{Model}}}
& \multirow{2}{*}{\textbf{Method}}
& \multicolumn{1}{c}{\textbf{Knowledge Search}}
& \multicolumn{1}{c}{\textbf{Agentic Reasoning}}
& \multicolumn{1}{c}{\textbf{Mathematical Reasoning}}
& \multicolumn{1}{c}{\textbf{Scientific Reasoning}}
& \multirow{2}{*}{\textbf{Average}} \\
\cmidrule(lr){3-3}
\cmidrule(lr){4-4}
\cmidrule(lr){5-5}
\cmidrule(lr){6-6}
& & \textbf{Bamboogle}
& \textbf{GAIA}
& \textbf{AIME25}
& \textbf{GPQA}
& \\
\midrule

\multirow{5}{*}{Qwen3.8-27B}
& \fullkvcell{Vanilla}
& \fullkvcell{87.20}
& \fullkvcell{58.25}
& \fullkvcell{96.67}
& \fullkvcell{77.78}
& \fullkvcell{79.98} \\
\cdashline{2-7}

& KIVI       & 90.40 & 58.25 & 90.00 & 81.82 & 80.12 \\
& TurboQuant & 88.80 & 58.25 & 93.33 & 81.82 & 80.55 \\
& PM-KVQ     & 88.80 & 56.31 & 86.67 & 77.78 & 77.39 \\
\rowcolor{softcyan}
& \textbf{\method{}} & 91.20 &57.28 & 93.33 & 81.82 & \textbf{80.91} \\

\midrule

\multirow{5}{*}{Gemma-4-31B-it}
& \fullkvcell{Vanilla}
& \fullkvcell{69.60}
& \fullkvcell{29.13}
& \fullkvcell{93.33}
& \fullkvcell{81.82}
& \fullkvcell{68.47} \\
\cdashline{2-7}

& KIVI       & 69.60 & 28.16 & 90.00 & 80.81 & 67.14 \\
& TurboQuant & 68.00 & 27.18 & 90.00 & 85.86 & 67.76 \\
& PM-KVQ     & 61.60 & 32.04 & 90.00 & 84.85 & 67.12 \\
\rowcolor{softcyan}
& \textbf{\method{}} &70.40 & 28.16 & 90.00 & 84.85 & \textbf{68.35} \\

\bottomrule
\end{tabular}%
}
\end{table}

\paragraph{Benchmarks.}
We evaluate four benchmarks spanning four types of reasoning tasks: (1) \textit{knowledge-\allowbreak intensive search}, including Bamboogle~\citep{press2022compositionality}; (2) \textit{agentic reasoning}, including GAIA~\citep{gaia}; (3) \textit{mathematical reasoning}, including AIME25~\citep{aime25}; and (4) \textit{scientific reasoning}, represented by GPQA~\citep{gpqa}. Together, these benchmarks cover knowledge retrieval, multi-step agent interaction, mathematical problem solving, and challenging scientific reasoning.

\paragraph{Implementation Details.}
All experiments are run on a server equipped with 8$\times$ NVIDIA H20 96GB GPUs, and our implementation is built on SGLang~\citep{sglang}. For a fair comparison, we evaluate all methods under average KV-cache bit-width budgets of 2, 3, and 4 bits. Each method follows its original bit-allocation strategy while matching the corresponding average bit-width budget. Unless otherwise specified, \method{} uses a token block size of $P=128$, a quantization group size of $G=32$, a native-precision recent window of $R=128$ tokens, an observation window of $W=32$ tokens, and a sensitivity refresh interval of $J=128$ newly generated tokens.

\subsection{Main Results}

\paragraph{Comparison with quantization baselines.}
Table~\ref{tab:main_results} reports the results across four benchmarks and two model backbones under an average 3-bit KV-cache bit-width budget. On Qwen3.8-27B, \method{} achieves the highest average accuracy among all quantization methods, reaching 80.91\% and outperforming the strongest baseline, TurboQuant, by 0.36\%. On Gemma-4-31B-it, \method{} achieves an average accuracy of 68.35\%, outperforming the strongest quantization baseline, TurboQuant, by 0.59\%. Furthermore, \method{} remains competitive with the unquantized Vanilla reference, demonstrating the effectiveness of coordinated mixed-precision KV-cache quantization across different model architectures.

\begin{figure}[t]
    \centering
    \includegraphics[width=\textwidth]{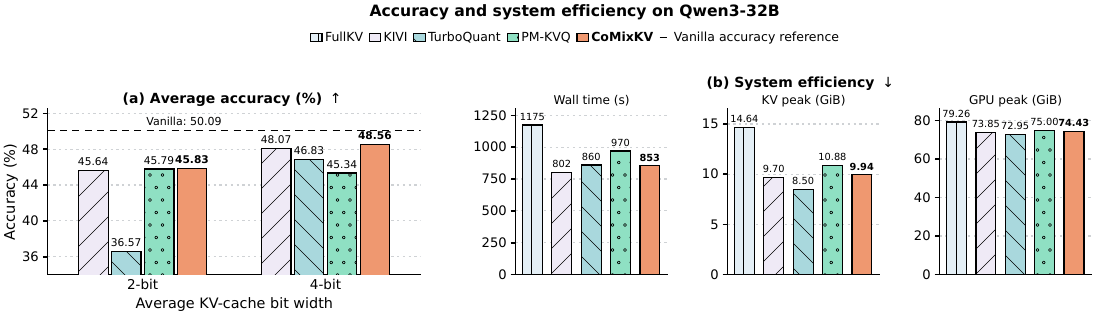}
    \caption{\textbf{Accuracy and system efficiency on Qwen3-32B.}
    (a) Average accuracy across four benchmarks under average
    2-bit and 4-bit KV-cache bit-width budgets; the dashed line denotes
    the unquantized 16-bit Vanilla reference.
    (b) System efficiency on the controlled long-context agent workload,
    including batch wall time, physical KV-cache peak, and total GPU-memory
    peak. Lower is better for all three efficiency metrics.}
    \label{fig:accuracy_memory_latency}
    \vspace{-1em}
\end{figure}

\paragraph{Different Bit-width budgets.}
Figure~\ref{fig:accuracy_memory_latency}(a) summarizes the average accuracy of Qwen3-32B under average KV-cache bit-width budgets of 2 and 4 bits, while Table~\ref{tab:bitwidth_results} in Appendix~\ref{app:qwen3_bitwidth_results} reports the complete benchmark-level results. \method{} achieves the highest average accuracy among all quantization methods at both bit-width budgets. At 2 bits, \method{} obtains 45.83\% average accuracy, slightly outperforming the strongest baseline, PM-KVQ, by 0.04\%. At 4 bits, \method{} reaches 48.56\%, exceeding the strongest baseline, KIVI, by 0.49\%. These results show that \method{} remains effective across different KV-cache precision budgets.

\subsection{Memory and Latency Analysis}
\label{sec:memory_latency}

\paragraph{Controlled long-context agent workload.}
We evaluate system efficiency on Qwen3-32B using a controlled three-stage Manager--Analyst--Verifier workflow. Each call uses $32{,}768$ input tokens and generates $2{,}048$ output tokens, with five samples executed concurrently on one NVIDIA H20 96GB GPU. All methods use the same inputs and decoding configuration; quantized methods use the average 2-bit setting, while FullKV stores KV states in BF16.

\paragraph{Results.}
Figure~\ref{fig:accuracy_memory_latency}(b) shows that \method{} reduces batch wall time from $1174.59$\,s to $853.44$\,s, corresponding to a $1.38\times$ speedup over FullKV. It also reduces the KV-cache peak from $14.64$\,GiB to $9.94$\,GiB ($32.08\%$ reduction) and the total GPU-memory peak from $79.26$\,GiB to $74.43$\,GiB. These results indicate that \method{} substantially lowers KV-cache memory while improving end-to-end throughput under the same GPU-memory budget.

\subsection{Ablation Studies}

We further analyze the key design choices of \method{} in Appendix~\ref{app:sensitivity_hyperparameters} and Appendix~\ref{app:online_allocation_ablation}. For hyperparameter sensitivity, the default setting $(W,J)=(32,128)$ ties for the best observed accuracy while avoiding a larger observation window or more frequent sensitivity updates, providing a favorable trade-off between accuracy and online overhead. For the allocation strategy, the online sensitivity-based policy consistently performs best among the evaluated variants, outperforming both uniform allocation and fixed sensitivity. These results demonstrate that dynamically updating layer sensitivity during execution and reallocating KV-cache precision accordingly is an important component of \method{}.

\section{Conclusion}

In this paper, we propose \method{}, a mixed-precision KV-cache quantization method for efficient LLM-based multi-agent systems. Our method assigns bit widths to agent--layer cache blocks under a shared memory budget. We derive a sensitivity score from a local attention-error bound and compute it using recent queries, keys, and values, without a calibration dataset or additional model passes. Paired lower- and higher-bit layer choices preserve each invocation's target mean packed-code width and cover all three candidates. The selected widths apply only to new blocks, while previously quantized blocks remain unchanged. Periodic sensitivity measurements refresh layer assignments as agents' states change.

\section*{Generative AI Disclosure}

Generative AI tools were used to assist with research brainstorming and providing feedback on methodology and experimental design, literature retrieval and summarization, drafting and editing parts of the manuscript, translation, code editing, and improving the clarity of technical descriptions and scientific figures. Generative AI tools were also used to assist with mathematical derivations and the development and verification of arguments supporting mathematical claims. All methodological decisions, implementations, experimental results, mathematical derivations, proofs, and scientific claims were independently reviewed and verified by the authors. The authors made all final research decisions and take full responsibility for the content of this work.

\section*{Ethics Statement}

This work does not involve human subjects or private user data. We follow the ICLR Code of Ethics and do not identify any specific ethical concerns associated with the methods or experiments presented in this work.

\section*{Reproducibility Statement}

We provide detailed descriptions of the experimental settings, models, datasets, baselines, and evaluation protocols in the Experiments section. The code will be made publicly available upon acceptance of the paper.

\bibliography{references}

\begin{thebibliography}{64}
\providecommand{\natexlab}[1]{#1}
\providecommand{\url}[1]{\texttt{#1}}
\expandafter\ifx\csname urlstyle\endcsname\relax
  \providecommand{\doi}[1]{doi: #1}\else
  \providecommand{\doi}{doi: \begingroup \urlstyle{rm}\Url}\fi

\bibitem[Bian et~al.(2025)Bian, Wu, Li, Ma, and Zhuo]{aq_tokencake}
Zhuohang Bian, Feiyang Wu, Zhuoran Li, Teng Ma, and Youwei Zhuo.
\newblock Tokencake: A kv-cache-centric serving framework for llm-based multi-agent applications.
\newblock \emph{arXiv preprint arXiv:2510.18586}, 2025.

\bibitem[Bondy et~al.(1976)Bondy, Murty, et~al.]{bondy1976graph}
John~Adrian Bondy, Uppaluri Siva~Ramachandra Murty, et~al.
\newblock \emph{Graph theory with applications}, volume 290.
\newblock Macmillan London, 1976.

\bibitem[Cai et~al.(2025)Cai, Zhang, Gao, Liu, Li, Liu, Lu, Xiong, Dong, Hu, and Xiao]{pyramidkv}
Zefan Cai, Yichi Zhang, Bofei Gao, Yuliang Liu, Yucheng Li, Tianyu Liu, Keming Lu, Wayne Xiong, Yue Dong, Junjie Hu, and Wen Xiao.
\newblock Pyramid{KV}: Dynamic {KV} cache compression based on pyramidal information funneling.
\newblock In \emph{Second Conference on Language Modeling}, 2025.
\newblock URL \url{https://openreview.net/forum?id=ayi7qezU87}.

\bibitem[Chen et~al.(2024)Chen, Liu, Han, Zhang, and Liu]{chen2024survey}
Shuaihang Chen, Yuanxing Liu, Wei Han, Weinan Zhang, and Ting Liu.
\newblock A survey on llm-based multi-agent system: Recent advances and new frontiers in application.
\newblock \emph{arXiv preprint arXiv:2412.17481}, 2024.

\bibitem[Chitty-Venkata et~al.(2026)Chitty-Venkata, Ye, Raskar, Kougkas, Sun, Emani, Vishwanath, and Nicolae]{pagedeviction}
Krishna~Teja Chitty-Venkata, Jie Ye, Siddhisanket Raskar, Anthony Kougkas, Xian Sun, Murali Emani, Venkatram Vishwanath, and Bogdan Nicolae.
\newblock Pagedeviction: Structured block-wise kv cache pruning for efficient large language model inference.
\newblock In \emph{Findings of the Association for Computational Linguistics: EACL 2026}, pages 3207--3218, 2026.

\bibitem[Dong et~al.(2024)Dong, Cheng, Qin, and Wang]{qaq}
Shichen Dong, Wen Cheng, Jiayu Qin, and Wei Wang.
\newblock Qaq: Quality adaptive quantization for llm kv cache.
\newblock \emph{arXiv preprint arXiv:2403.04643}, 2024.

\bibitem[Feng et~al.(2025)Feng, Lv, Cao, Xie, and Zhou]{adakv}
Yuan Feng, Junlin Lv, Yukun Cao, Xike Xie, and S~Kevin Zhou.
\newblock Ada-kv: Optimizing kv cache eviction by adaptive budget allocation for efficient llm inference.
\newblock \emph{Advances in Neural Information Processing Systems}, 38:\penalty0 113152--113188, 2025.

\bibitem[Fourney et~al.(2024)Fourney, Bansal, Mozannar, Tan, Salinas, Erkang, Zhu, Niedtner, Proebsting, Bassman, Gerrits, Alber, Chang, Loynd, West, Dibia, Awadallah, Kamar, Hosn, and Amershi]{fourney2024magentic}
Adam Fourney, Gagan Bansal, Hussein Mozannar, Cheng Tan, Eduardo Salinas, Erkang, Zhu, Friederike Niedtner, Grace Proebsting, Griffin Bassman, Jack Gerrits, Jacob Alber, Peter Chang, Ricky Loynd, Robert West, Victor Dibia, Ahmed Awadallah, Ece Kamar, Rafah Hosn, and Saleema Amershi.
\newblock Magentic-one: A generalist multi-agent system for solving complex tasks, 2024.
\newblock URL \url{https://arxiv.org/abs/2411.04468}.

\bibitem[Gao et~al.(2024{\natexlab{a}})Gao, He, Sharma, Kang, Jevdjic, Deng, Yang, Yu, and Zuo]{gao2024cost}
Bin Gao, Zhuomin He, Puru Sharma, Qingxuan Kang, Djordje Jevdjic, Junbo Deng, Xingkun Yang, Zhou Yu, and Pengfei Zuo.
\newblock Cost-efficient large language model serving for multi-turn conversations with cachedattention.
\newblock In \emph{2024 USENIX annual technical conference (USENIX ATC 24)}, pages 111--126, 2024{\natexlab{a}}.

\bibitem[Gao et~al.(2024{\natexlab{b}})Gao, Li, Pan, Kuang, Ma, Qian, Wei, Zhang, Xie, Chen, et~al.]{agentscope}
Dawei Gao, Zitao Li, Xuchen Pan, Weirui Kuang, Zhijian Ma, Bingchen Qian, Fei Wei, Wenhao Zhang, Yuexiang Xie, Daoyuan Chen, et~al.
\newblock Agentscope: A flexible yet robust multi-agent platform.
\newblock \emph{arXiv preprint arXiv:2402.14034}, 2024{\natexlab{b}}.

\bibitem[Ge et~al.(2024)Ge, Zhang, Liu, Zhang, Han, and Gao]{ge2024model}
Suyu Ge, Yunan Zhang, Liyuan Liu, Minjia Zhang, Jiawei Han, and Jianfeng Gao.
\newblock Model tells you what to discard: Adaptive kv cache compression for llms.
\newblock In \emph{International Conference on Learning Representations}, volume 2024, pages 22975--22988, 2024.

\bibitem[Guo et~al.(2024)Guo, Chen, Wang, Chang, Pei, Chawla, Wiest, and Zhang]{guo2024large}
Taicheng Guo, Xiuying Chen, Yaqi Wang, Ruidi Chang, Shichao Pei, Nitesh~V. Chawla, Olaf Wiest, and Xiangliang Zhang.
\newblock Large language model based multi-agents: A survey of progress and challenges.
\newblock In Kate Larson, editor, \emph{Proceedings of the Thirty-Third International Joint Conference on Artificial Intelligence, {IJCAI-24}}, pages 8048--8057. International Joint Conferences on Artificial Intelligence Organization, 8 2024.
\newblock \doi{10.24963/ijcai.2024/890}.
\newblock URL \url{https://doi.org/10.24963/ijcai.2024/890}.
\newblock Survey Track.

\bibitem[Hariri et~al.(2026)Hariri, Luo, Chen, Zhang, Wang, Han, and Chaudhary]{hariri2026quantize}
Mohsen Hariri, Alan Luo, Weicong Chen, Tianyi Zhang, Qifan Wang, Xiaotian Han, and Vipin Chaudhary.
\newblock Quantize what counts: More for keys, less for values.
\newblock In \emph{Findings of the Association for Computational Linguistics: ACL 2026}, pages 26393--26420, 2026.

\bibitem[He et~al.(2025)He, Treude, and Lo]{he2025llm}
Junda He, Christoph Treude, and David Lo.
\newblock Llm-based multi-agent systems for software engineering: Literature review, vision, and the road ahead.
\newblock \emph{ACM Transactions on Software Engineering and Methodology}, 34\penalty0 (5):\penalty0 1--30, 2025.

\bibitem[He et~al.(2024)He, Zhang, Wu, Liu, Zhou, and Zhuang]{zipcache}
Yefei He, Luoming Zhang, Weijia Wu, Jing Liu, Hong Zhou, and Bohan Zhuang.
\newblock Zipcache: Accurate and efficient kv cache quantization with salient token identification.
\newblock \emph{Advances in Neural Information Processing Systems}, 37:\penalty0 68287--68307, 2024.

\bibitem[Honavar and GVSL(2026)]{aq_qkvshare}
Pratik Honavar and Tejpratap GVSL.
\newblock Qkvshare: Quantized kv-cache handoff for multi-agent on-device llms.
\newblock \emph{arXiv preprint arXiv:2605.03884}, 2026.

\bibitem[Hong et~al.(2024)Hong, Zhuge, Chen, Zheng, Cheng, Wang, Zhang, Yau, Lin, Zhou, et~al.]{metagpt}
Sirui Hong, Mingchen Zhuge, Jonathan Chen, Xiawu Zheng, Yuheng Cheng, Jinlin Wang, Ceyao Zhang, Steven Yau, Zijuan Lin, Liyang Zhou, et~al.
\newblock Metagpt: Meta programming for a multi-agent collaborative framework.
\newblock In \emph{International Conference on Learning Representations}, volume 2024, pages 23247--23275, 2024.

\bibitem[Hooper et~al.(2024)Hooper, Kim, Mohammadzadeh, Mahoney, Shao, Keutzer, and Gholami]{kvquant}
Coleman Hooper, Sehoon Kim, Hiva Mohammadzadeh, Michael~W Mahoney, Yakun~S Shao, Kurt Keutzer, and Amir Gholami.
\newblock Kvquant: Towards 10 million context length llm inference with kv cache quantization.
\newblock \emph{Advances in Neural Information Processing Systems}, 37:\penalty0 1270--1303, 2024.

\bibitem[Hu et~al.(2026)Hu, Jia, Wang, Li, Yang, Hao, Li, and Liao]{take}
Long Hu, Nan Jia, Rui Wang, Jiahui Li, Qingyi Yang, Yixue Hao, Xianzhi Li, and Xiaofei Liao.
\newblock {TAKE}: Task-aware chunked {KV} cache eviction for efficient long-context {LLM} prefill, 2026.
\newblock URL \url{https://openreview.net/forum?id=kMLfUshPwo}.

\bibitem[Jin et~al.(2025)Jin, Liu, Zhang, and Mao]{jin2025compute}
Shuowei Jin, Xueshen Liu, Qingzhao Zhang, and Zhuoqing Mao.
\newblock Compute or load {KV} cache? why not both?
\newblock In \emph{Forty-second International Conference on Machine Learning}, 2025.
\newblock URL \url{https://openreview.net/forum?id=WOyOtaO6lQ}.

\bibitem[Kang et~al.(2024)Kang, Zhang, Kundu, Jeong, Liu, Krishna, and Zhao]{gear}
Hao Kang, Qingru Zhang, Souvik Kundu, Geonhwa Jeong, Zaoxing Liu, Tushar Krishna, and Tuo Zhao.
\newblock Gear: An efficient kv cache compression recipe for near-lossless generative inference of llm.
\newblock \emph{arXiv preprint arXiv:2403.05527}, 2024.

\bibitem[Kriuk and Ng(2026)]{aq_qkvcomm}
Boris Kriuk and Logic Ng.
\newblock Q-kvcomm: Efficient multi-agent communication via adaptive kv cache compression.
\newblock In \emph{2026 Second International Conference on Multi-Agent Systems for Collaborative Intelligence (ICMSCI)}, pages 15--21. IEEE, 2026.

\bibitem[Kwon et~al.(2023)Kwon, Li, Zhuang, Sheng, Zheng, Yu, Gonzalez, Zhang, and Stoica]{vllm}
Woosuk Kwon, Zhuohan Li, Siyuan Zhuang, Ying Sheng, Lianmin Zheng, Cody~Hao Yu, Joseph Gonzalez, Hao Zhang, and Ion Stoica.
\newblock Efficient memory management for large language model serving with pagedattention.
\newblock In \emph{Proceedings of the 29th symposium on operating systems principles}, pages 611--626, 2023.

\bibitem[Li et~al.(2023)Li, Hammoud, Itani, Khizbullin, and Ghanem]{li2023camel}
Guohao Li, Hasan Hammoud, Hani Itani, Dmitrii Khizbullin, and Bernard Ghanem.
\newblock Camel: Communicative agents for" mind" exploration of large language model society.
\newblock \emph{Advances in neural information processing systems}, 36:\penalty0 51991--52008, 2023.

\bibitem[Li et~al.(2024{\natexlab{a}})Li, Li, Tian, Tang, Xu, Chen, Hu, Dong, Li, and Chen]{li2024survey}
Haoyang Li, Yiming Li, Anxin Tian, Tianhao Tang, Zhanchao Xu, Xuejia Chen, Nicole Hu, Wei Dong, Qing Li, and Lei Chen.
\newblock A survey on large language model acceleration based on kv cache management.
\newblock \emph{arXiv preprint arXiv:2412.19442}, 2024{\natexlab{a}}.

\bibitem[Li et~al.(2024{\natexlab{b}})Li, Huang, Yang, Venkitesh, Locatelli, Ye, Cai, Lewis, and Chen]{snapkv}
Yuhong Li, Yingbing Huang, Bowen Yang, Bharat Venkitesh, Acyr Locatelli, Hanchen Ye, Tianle Cai, Patrick Lewis, and Deming Chen.
\newblock Snapkv: Llm knows what you are looking for before generation.
\newblock \emph{Advances in Neural Information Processing Systems}, 37:\penalty0 22947--22970, 2024{\natexlab{b}}.

\bibitem[Liu et~al.(2024{\natexlab{a}})Liu, Liu, Pan, He, Haffari, and Zhuang]{minicache}
Akide Liu, Jing Liu, Zizheng Pan, Yefei He, Gholamreza Haffari, and Bohan Zhuang.
\newblock Minicache: Kv cache compression in depth dimension for large language models.
\newblock \emph{Advances in Neural Information Processing Systems}, 37:\penalty0 139997--140031, 2024{\natexlab{a}}.

\bibitem[Liu et~al.(2026)Liu, Li, Yang, Zhao, Zhou, Song, Dai, Yan, Yang, and Wang]{liu2026pm}
Tengxuan Liu, Shiyao Li, Jiayi Yang, Tianchen Zhao, Feng Zhou, Xiaohui Song, Guohao Dai, Shengen Yan, Huazhong Yang, and Yu~Wang.
\newblock Pm-kvq: Progressive mixed-precision kv cache quantization for long-cot llms.
\newblock In \emph{International Conference on Learning Representations}, volume 2026, pages 107881--107900, 2026.

\bibitem[Liu et~al.(2023)Liu, Desai, Liao, Wang, Xie, Xu, Kyrillidis, and Shrivastava]{scissorhands}
Zichang Liu, Aditya Desai, Fangshuo Liao, Weitao Wang, Victor Xie, Zhaozhuo Xu, Anastasios Kyrillidis, and Anshumali Shrivastava.
\newblock Scissorhands: Exploiting the persistence of importance hypothesis for llm kv cache compression at test time.
\newblock \emph{Advances in Neural Information Processing Systems}, 36:\penalty0 52342--52364, 2023.

\bibitem[Liu et~al.(2024{\natexlab{b}})Liu, Zhang, Li, Liu, and Yang]{dylan}
Zijun Liu, Yanzhe Zhang, Peng Li, Yang Liu, and Diyi Yang.
\newblock A dynamic llm-powered agent network for task-oriented agent collaboration.
\newblock In \emph{First Conference on Language Modeling}, 2024{\natexlab{b}}.
\newblock URL \url{https://openreview.net/forum?id=XII0Wp1XA9}.

\bibitem[Liu et~al.(2024{\natexlab{c}})Liu, Yuan, Jin, Zhong, Xu, Braverman, Chen, and Hu]{kivi}
Zirui Liu, Jiayi Yuan, Hongye Jin, Shaochen Zhong, Zhaozhuo Xu, Vladimir Braverman, Beidi Chen, and Xia Hu.
\newblock Kivi: A tuning-free asymmetric 2bit quantization for kv cache.
\newblock In \emph{International Conference on Machine Learning}, pages 32332--32344. PMLR, 2024{\natexlab{c}}.

\bibitem[Masterman et~al.(2024)Masterman, Besen, Sawtell, and Chao]{masterman2024landscape}
Tula Masterman, Sandi Besen, Mason Sawtell, and Alex Chao.
\newblock The landscape of emerging ai agent architectures for reasoning, planning, and tool calling: A survey.
\newblock \emph{arXiv preprint arXiv:2404.11584}, 2024.

\bibitem[Mialon et~al.(2024)Mialon, Fourrier, Wolf, LeCun, and Scialom]{gaia}
Gr{\'e}goire Mialon, Cl{\'e}mentine Fourrier, Thomas Wolf, Yann LeCun, and Thomas Scialom.
\newblock Gaia: a benchmark for general ai assistants.
\newblock In \emph{International Conference on Learning Representations}, volume 2024, pages 9025--9049, 2024.

\bibitem[Pan et~al.(2025)Pan, PATEL, Shen, Hu, Guan, Li, Qin, Wang, and Ding]{pan2025kvflow}
Zaifeng Pan, AJJKUMAR~DAHYALAL PATEL, Yipeng Shen, Zhengding Hu, Yue Guan, Wan-Lu Li, Lianhui Qin, Yida Wang, and Yufei Ding.
\newblock Kvflow: Efficient prefix caching for accelerating llm-based multi-agent workflows.
\newblock \emph{Advances in Neural Information Processing Systems}, 38:\penalty0 126246--126265, 2025.

\bibitem[Patel and Joshi(2026)]{aq_polykv}
Ishan Patel and Ishan Joshi.
\newblock Polykv: A shared asymmetrically-compressed kv cache pool for multi-agent llm inference.
\newblock \emph{arXiv preprint arXiv:2604.24971}, 2026.

\bibitem[Press et~al.(2023)Press, Zhang, Min, Schmidt, Smith, and Lewis]{press2022compositionality}
Ofir Press, Muru Zhang, Sewon Min, Ludwig Schmidt, Noah Smith, and Mike Lewis.
\newblock Measuring and narrowing the compositionality gap in language models.
\newblock In Houda Bouamor, Juan Pino, and Kalika Bali, editors, \emph{Findings of the Association for Computational Linguistics: EMNLP 2023}, pages 5687--5711, Singapore, December 2023. Association for Computational Linguistics.
\newblock \doi{10.18653/v1/2023.findings-emnlp.378}.
\newblock URL \url{https://aclanthology.org/2023.findings-emnlp.378/}.

\bibitem[Qian et~al.(2024)Qian, Liu, Liu, Chen, Dang, Li, Yang, Chen, Su, Cong, et~al.]{qian2024chatdev}
Chen Qian, Wei Liu, Hongzhang Liu, Nuo Chen, Yufan Dang, Jiahao Li, Cheng Yang, Weize Chen, Yusheng Su, Xin Cong, et~al.
\newblock Chatdev: Communicative agents for software development.
\newblock In \emph{Proceedings of the 62nd annual meeting of the association for computational linguistics (volume 1: Long papers)}, pages 15174--15186, 2024.

\bibitem[{Qwen Team}(2026)]{qwen38_27b}
{Qwen Team}.
\newblock {Qwen3.8-27B} model card.
\newblock Hugging Face, 2026.
\newblock URL \url{https://huggingface.co/Qwen/Qwen3.8-27B}.

\bibitem[Rein et~al.(2024)Rein, Hou, Stickland, Petty, Pang, Dirani, Michael, and Bowman]{gpqa}
David Rein, Betty~Li Hou, Asa~Cooper Stickland, Jackson Petty, Richard~Yuanzhe Pang, Julien Dirani, Julian Michael, and Samuel~R. Bowman.
\newblock Gpqa: A graduate-level google-proof q\&a benchmark.
\newblock In \emph{First Conference on Language Modeling}, 2024.
\newblock URL \url{https://openreview.net/forum?id=Ti67584b98}.

\bibitem[Shen et~al.(2026)Shen, Wu, Zhao, and Mullins]{shen2026triaxialkv}
Hanzhang Shen, Haoran Wu, Yiren Zhao, and Robert Mullins.
\newblock Triaxialkv: Toward extreme low-precision kv-cache quantization for agentic inference tasks.
\newblock \emph{arXiv preprint arXiv:2605.17170}, 2026.

\bibitem[Tang et~al.(2024)Tang, Zhao, Zhu, Xiao, Kasikci, and Han]{quest}
Jiaming Tang, Yilong Zhao, Kan Zhu, Guangxuan Xiao, Baris Kasikci, and Song Han.
\newblock Quest: Query-aware sparsity for efficient long-context llm inference.
\newblock In \emph{International Conference on Machine Learning}, pages 47901--47911. PMLR, 2024.

\bibitem[Team et~al.(2026)Team, Abd, Aggarwal, Algayres, Andreev, Bachem, Ballantyne, Brick, C{\u{a}}rbune, Casbon, et~al.]{team2026gemma}
Gemma Team, Sherif~El Abd, Vaibhav Aggarwal, Robin Algayres, Alek Andreev, Olivier Bachem, Ian Ballantyne, Cormac Brick, Victor C{\u{a}}rbune, Michelle Casbon, et~al.
\newblock Gemma 4 technical report.
\newblock \emph{arXiv preprint arXiv:2607.02770}, 2026.

\bibitem[Wang et~al.(2025)Wang, Han, Wei, Shen, Zhang, Fang, Chen, Yu, and Chen]{wang2025kvcache}
Jiahao Wang, Jinbo Han, Xingda Wei, Sijie Shen, Dingyan Zhang, Chenguang Fang, Rong Chen, Wenyuan Yu, and Haibo Chen.
\newblock $\{$KVCache$\}$ cache in the wild: Characterizing and optimizing $\{$KVCache$\}$ cache at a large cloud provider.
\newblock In \emph{2025 USENIX Annual Technical Conference (USENIX ATC 25)}, pages 465--482, 2025.

\bibitem[Wu et~al.(2024)Wu, Bansal, Zhang, Wu, Li, Zhu, Jiang, Zhang, Zhang, Liu, Awadallah, White, Burger, and Wang]{wu2024autogen}
Qingyun Wu, Gagan Bansal, Jieyu Zhang, Yiran Wu, Beibin Li, Erkang Zhu, Li~Jiang, Xiaoyun Zhang, Shaokun Zhang, Jiale Liu, Ahmed~Hassan Awadallah, Ryen~W White, Doug Burger, and Chi Wang.
\newblock Autogen: Enabling next-gen {LLM} applications via multi-agent conversations.
\newblock In \emph{First Conference on Language Modeling}, 2024.
\newblock URL \url{https://openreview.net/forum?id=BAakY1hNKS}.

\bibitem[Wu et~al.(2025)Wu, Lv, feng, zhang, Zhang, Yin, Lin, and Yan]{wu2026polarquant}
Songhao Wu, Ang Lv, xiao feng, Yufei zhang, Xun Zhang, Guojun Yin, Wei Lin, and Rui Yan.
\newblock Polarquant: Leveraging polar transformation for key cache quantization and decoding acceleration.
\newblock In D.~Belgrave, C.~Zhang, H.~Lin, R.~Pascanu, P.~Koniusz, M.~Ghassemi, and N.~Chen, editors, \emph{Advances in Neural Information Processing Systems}, volume 38, Main Conference, pages 50584--50607. Curran Associates, Inc., 2025.
\newblock \doi{10.52202/085713-1688}.
\newblock URL \url{https://proceedings.neurips.cc/paper_files/paper/2025/file/48aa1c882f99f5cdd3c76f7c0ede44a7-Paper-Conference.pdf}.

\bibitem[Yang et~al.(2025)Yang, Li, Yang, Zhang, Hui, Zheng, Yu, Gao, Huang, Lv, et~al.]{yang2025qwen3}
An~Yang, Anfeng Li, Baosong Yang, Beichen Zhang, Binyuan Hui, Bo~Zheng, Bowen Yu, Chang Gao, Chengen Huang, Chenxu Lv, et~al.
\newblock Qwen3 technical report.
\newblock \emph{arXiv preprint arXiv:2505.09388}, 2025.

\bibitem[Yang et~al.(2024)Yang, Peng, Wang, Wen, and Zhang]{yang2024llm}
Yingxuan Yang, Qiuying Peng, Jun Wang, Ying Wen, and Weinan Zhang.
\newblock Llm-based multi-agent systems: Techniques and business perspectives.
\newblock \emph{arXiv preprint arXiv:2411.14033}, 2024.

\bibitem[Yao et~al.(2023)Yao, Zhao, Yu, Du, Shafran, Narasimhan, and Cao]{react}
Shunyu Yao, Jeffrey Zhao, Dian Yu, Nan Du, Izhak Shafran, Karthik~R Narasimhan, and Yuan Cao.
\newblock React: Synergizing reasoning and acting in language models.
\newblock In \emph{The Eleventh International Conference on Learning Representations}, 2023.
\newblock URL \url{https://openreview.net/forum?id=WE_vluYUL-X}.

\bibitem[Ye et~al.(2025)Ye, Gao, Ma, Wang, Fu, Chung, Lin, Liu, Zhang, Zhuo, et~al.]{ye2026kvcomm}
Hancheng Ye, Zhengqi Gao, Mingyuan Ma, Qinsi Wang, Yuzhe Fu, Ming-Yu Chung, Yueqian Lin, Zhijian Liu, Jianyi Zhang, Danyang Zhuo, et~al.
\newblock Kvcomm: Online cross-context kv-cache communication for efficient llm-based multi-agent systems.
\newblock \emph{Advances in Neural Information Processing Systems}, 38:\penalty0 17882--17928, 2025.

\bibitem[Yu et~al.(2025)Yu, Lin, and Li]{yu2025stateful}
Lingfan Yu, Jinkun Lin, and Jinyang Li.
\newblock Stateful large language model serving with pensieve.
\newblock In \emph{Proceedings of the Twentieth European Conference on Computer Systems}, pages 144--158, 2025.

\bibitem[Zandieh et~al.(2026)Zandieh, Daliri, Hadian, and Mirrokni]{zandieh2026turboquant}
Amir Zandieh, Majid Daliri, Majid Hadian, and Vahab Mirrokni.
\newblock Turboquant: Online vector quantization with near-optimal distortion rate.
\newblock In \emph{International Conference on Learning Representations}, volume 2026, pages 56418--56439, 2026.

\bibitem[Zhang et~al.(2025{\natexlab{a}})Zhang, Yue, Li, Yun, Wan, Wang, Cheng, Yu, and Chen]{agentprune}
Guibin Zhang, Yanwei Yue, Zhixun Li, Sukwon Yun, Guancheng Wan, Kun Wang, Dawei Cheng, Jeffrey Yu, and Tianlong Chen.
\newblock Cut the crap: An economical communication pipeline for llm-based multi-agent systems.
\newblock In \emph{International Conference on Learning Representations}, volume 2025, pages 75389--75428, 2025{\natexlab{a}}.

\bibitem[Zhang et~al.(2025{\natexlab{b}})Zhang, Ji, Chen, Fu, Miao, Nie, Chen, and Cui]{pqcache}
Hailin Zhang, Xiaodong Ji, Yilin Chen, Fangcheng Fu, Xupeng Miao, Xiaonan Nie, Weipeng Chen, and Bin Cui.
\newblock Pqcache: Product quantization-based kvcache for long context llm inference.
\newblock \emph{Proceedings of the ACM on Management of Data}, 3\penalty0 (3):\penalty0 1--30, 2025{\natexlab{b}}.

\bibitem[Zhang et~al.(2026{\natexlab{a}})Zhang, Li, Feng, Yuan, Wang, Zhang, Shi, Tan, Pan, Hu, and Li]{zhang2026patternkv}
Ji~Zhang, Yiwei Li, Shaoxiong Feng, Peiwen Yuan, Xinglin Wang, Yueqi Zhang, Jiayi Shi, Chuyi Tan, Boyuan Pan, Yao Hu, and Kan Li.
\newblock Pattern{KV}: Flattening {KV} representation expands quantization headroom.
\newblock In \emph{Forty-third International Conference on Machine Learning}, 2026{\natexlab{a}}.
\newblock URL \url{https://openreview.net/forum?id=ozV2cpEgnZ}.

\bibitem[Zhang et~al.(2025{\natexlab{c}})Zhang, Zhu, Song, Wu, Kuang, Li, Shang, Liu, and Li]{zhang2025more}
Jiebin Zhang, Dawei Zhu, Yifan Song, Wenhao Wu, Chuqiao Kuang, Xiaoguang Li, Lifeng Shang, Qun Liu, and Sujian Li.
\newblock More tokens, lower precision: Towards the optimal token-precision trade-off in kv cache compression.
\newblock In \emph{EMNLP (Findings)}, pages 8092--8105, 2025{\natexlab{c}}.

\bibitem[Zhang et~al.(2026{\natexlab{b}})Zhang, Zeng, Peng, Zhuang, and Chen]{aq_mixkvq}
Tao Zhang, Ziqian Zeng, Hao Peng, Huiping Zhuang, and Cen Chen.
\newblock Mixkvq: Query-aware mixed-precision kv cache quantization for long-context reasoning.
\newblock In \emph{Proceedings of the 64th Annual Meeting of the Association for Computational Linguistics (Volume 1: Long Papers)}, pages 7189--7204, 2026{\natexlab{b}}.

\bibitem[Zhang et~al.(2026{\natexlab{c}})Zhang, Cui, Wang, Qiu, Li, Han, Huang, Qiu, Zhu, and He]{zhang2026verified}
Xing Zhang, Yanwei Cui, Guanghui Wang, Wei Qiu, Ziyuan Li, Fangwei Han, Yajing Huang, Hengzhi Qiu, Bing Zhu, and Peiyang He.
\newblock Verified multi-agent orchestration: A plan-execute-verify-replan framework for complex query resolution.
\newblock \emph{arXiv preprint arXiv:2603.11445}, 2026{\natexlab{c}}.

\bibitem[Zhang et~al.(2025{\natexlab{d}})Zhang, Zeng, Li, Hu, Han, and Zuo]{zhang2025tool}
Yabo Zhang, Yihan Zeng, Qingyun Li, Zhen Hu, Kavin Han, and Wangmeng Zuo.
\newblock Tool-r1: Sample-efficient reinforcement learning for agentic tool use.
\newblock \emph{arXiv preprint arXiv:2509.12867}, 2025{\natexlab{d}}.

\bibitem[Zhang and Team(2025)]{aime25}
Yifan Zhang and Math-AI Team.
\newblock American invitational mathematics examination (aime) 2025, 2025.
\newblock URL \url{https://huggingface.co/datasets/math-ai/aime25}.

\bibitem[Zhang et~al.(2024)Zhang, Du, Luo, Zhong, Zhang, Liu, and Ji]{cam}
Yuxin Zhang, Yuxuan Du, Gen Luo, Yunshan Zhong, Zhenyu Zhang, Shiwei Liu, and Rongrong Ji.
\newblock Cam: Cache merging for memory-efficient llms inference.
\newblock In \emph{International Conference on Machine Learning}, pages 58840--58850. PMLR, 2024.

\bibitem[Zhang et~al.(2023)Zhang, Sheng, Zhou, Chen, Zheng, Cai, Song, Tian, R{\'e}, Barrett, et~al.]{h2o}
Zhenyu Zhang, Ying Sheng, Tianyi Zhou, Tianlong Chen, Lianmin Zheng, Ruisi Cai, Zhao Song, Yuandong Tian, Christopher R{\'e}, Clark Barrett, et~al.
\newblock H2o: Heavy-hitter oracle for efficient generative inference of large language models.
\newblock \emph{Advances in Neural Information Processing Systems}, 36:\penalty0 34661--34710, 2023.

\bibitem[Zheng et~al.(2024)Zheng, Yin, Xie, Sun, Huang, Yu, Cao, Kozyrakis, Stoica, Gonzalez, et~al.]{sglang}
Lianmin Zheng, Liangsheng Yin, Zhiqiang Xie, Chuyue Sun, Jeff Huang, Cody~H Yu, Shiyi Cao, Christos Kozyrakis, Ion Stoica, Joseph~E Gonzalez, et~al.
\newblock Sglang: Efficient execution of structured language model programs.
\newblock \emph{Advances in neural information processing systems}, 37:\penalty0 62557--62583, 2024.

\bibitem[Zhuge et~al.(2024)Zhuge, Wang, Kirsch, Faccio, Khizbullin, and Schmidhuber]{gptswarm}
Mingchen Zhuge, Wenyi Wang, Louis Kirsch, Francesco Faccio, Dmitrii Khizbullin, and J\"{u}rgen Schmidhuber.
\newblock Gptswarm: Language agents as optimizable graphs.
\newblock In Ruslan Salakhutdinov, Zico Kolter, Katherine Heller, Adrian Weller, Nuria Oliver, Jonathan Scarlett, and Felix Berkenkamp, editors, \emph{Proceedings of the 41st International Conference on Machine Learning}, volume 235 of \emph{Proceedings of Machine Learning Research}, pages 62743--62767. PMLR, 21--27 Jul 2024.
\newblock URL \url{https://proceedings.mlr.press/v235/zhuge24a.html}.

\bibitem[Zuo et~al.(2026)Zuo, Zhou, Cong, Xi, and Leung]{zuo2026ratequant}
Fei Zuo, Zikang Zhou, Hao Cong, Xiaoyan Xi, and Ho~Fai Leung.
\newblock Ratequant: Optimal mixed-precision kv cache quantization via rate-distortion theory.
\newblock \emph{arXiv preprint arXiv:2605.06675}, 2026.

\end{thebibliography}
\bibliographystyle{plainnat}

\appendix

\startcontents
\printcontents{}{1}{\section*{Appendix}}

\section{Additional Qwen3-32B Results at 2 and 4 Bits}
\label{app:qwen3_bitwidth_results}

Table~\ref{tab:bitwidth_results} reports supplementary Qwen3-32B accuracy results under average KV-cache bit-width budgets of 2 and 4 bits on the four benchmarks. The unquantized Vanilla reference is repeated for comparison.

\begin{table}[htbp]
\centering
\captionsetup{skip=2pt}
\fontsize{9.0}{9.3}\selectfont
\setlength{\tabcolsep}{1.8pt}
\renewcommand{\arraystretch}{1.15}

\caption{Accuracy (\%) of Qwen3-32B at average KV-cache bit-width budgets of 2 and 4 bits, with the unquantized 16-bit Vanilla reference.}
\label{tab:bitwidth_results}

\resizebox{\textwidth}{!}{%
\begin{tabular}{@{}llccccc@{}}
\toprule
\multirow{2}{*}{\shortstack[l]{\textbf{Bits}}}
& \multirow{2}{*}{\textbf{Method}}
& \multicolumn{1}{c}{\textbf{Knowledge Search}}
& \multicolumn{1}{c}{\textbf{Agentic Reasoning}}
& \multicolumn{1}{c}{\textbf{Mathematical Reasoning}}
& \multicolumn{1}{c}{\textbf{Scientific Reasoning}}
& \multirow{2}{*}{\textbf{Average}} \\
\cmidrule(lr){3-3}
\cmidrule(lr){4-4}
\cmidrule(lr){5-5}
\cmidrule(lr){6-6}
& & \textbf{Bamboogle}
& \textbf{GAIA}
& \textbf{AIME25}
& \textbf{GPQA}
& \\
\midrule

\fullkvcell{16}
& \fullkvcell{Vanilla}
& \fullkvcell{81.60}
& \fullkvcell{28.16}
& \fullkvcell{30.00}
& \fullkvcell{60.61}
& \fullkvcell{50.09} \\

\midrule

\multirow{4}{*}{2}
& KIVI
& 71.20 & 26.21 & 36.67 & 48.48 & 45.64 \\

& TurboQuant
& 53.60 & 21.36 & 23.33 & 47.98 & 36.57 \\

& PM-KVQ
& 71.20 & 24.27 & 36.67 & 51.01 & 45.79 \\

\rowcolor{softcyan}
& \textbf{\method{}}
& 69.60 &25.24 &40.00& 48.48 & \textbf{45.83} \\

\midrule

\multirow{4}{*}{4}
& KIVI
& 75.20 & 32.04 & 30.00 & 55.05 & 48.07 \\

& TurboQuant
& 73.60 & 28.16 & 30.00 & 55.56 & 46.83 \\

& PM-KVQ
& 76.00 & 28.16 & 26.67 & 50.51 & 45.34 \\

\rowcolor{softcyan}
& \textbf{\method{}}
&77.60 & 31.07 & 30.00 & 55.56 & \textbf{48.56} \\

\bottomrule
\end{tabular}%
}
\end{table}

\section{Notation and Runtime Workflow}
\label{app:runtime_workflow}

This section expands the execution graph in Section~\ref{sec:graph_formulation} and connects its agent states to the invocation-\allowbreak level caches in Section~\ref{sec:graph_conditioned_quantization}. We retain $i$ for an agent's role and $a$ for one invocation of that role; repeated invocations receive different identifiers even when they use the same role. The index $t$ denotes an agent step within an invocation, whereas $r$ denotes a cache-allocation round. Table~\ref{tab:appendix_notation} summarizes the notation. Bold lowercase symbols in these three sections represent serialized text sequences, and $\Vert$ concatenates them in order.

\begin{table}[htbp]
\centering
\caption{Agent-side notation and its connection to the main-text formulation.}
\label{tab:appendix_notation}
\small
\setlength{\tabcolsep}{5pt}
\renewcommand{\arraystretch}{1.12}
\begin{tabularx}{\linewidth}{@{}lX@{}}
\toprule
\textbf{Notation} & \textbf{Definition} \\
\midrule
$\boldsymbol{q}$ & Original task query in Eq.~(\ref{eq:agent_graph}); not the numerical attention query in Eq.~(\ref{eq:aq_attention}). \\
$i\in\{\mathrm{m},\mathrm{r},\mathrm{p},\mathrm{v}\}$ & Fixed role labels for manager, researcher, analyst, and verifier. The upright label $\mathrm{r}$ denotes the researcher, not allocation round $r$. \\
$a$ & One invocation of role $i$, owning the history $\boldsymbol{h}_a$ and cache $\mathcal{KV}_a$ defined in the main text. \\
$v_a^{(t)}$ & Invocation-indexed refinement of $v_i^{(t)}$, distinguishing separate runs of the same role. \\
$\boldsymbol{h}_a^{(t)}$ & The local history $\boldsymbol{h}_a$ before step $t$ of invocation $a$. \\
$\boldsymbol{y}_a^{(t)}$ & Serialized model output at that step, including its action, refining $\boldsymbol{y}_i^{(t)}$ in Eq.~(\ref{eq:agent_communication}). \\
$\boldsymbol{x}_a^{\mathrm{obs},(t)}$ & Text returned by the tool or managed-worker call issued at that step; not an attention-output vector. \\
$\mathcal{A}_r$ & Invocations whose caches are resident at allocation round $r$, including paused invocations; not the configured set of roles. \\
$\mathcal{E}_{\boldsymbol{q}}^{\mathrm{T}},\mathcal{E}_{\boldsymbol{q}}^{\mathrm{S}}$ & Temporal and spatial edge sets from Eq.~(\ref{eq:agent_graph}). \\
$\widehat{\boldsymbol{y}},\boldsymbol{z}$ & The manager's candidate answer and the final prediction accepted by the runtime. \\
$\boldsymbol{x}^{\mathrm{rep}},\boldsymbol{x}^{\mathrm{rec}}$ & Serialized worker evidence and additional recovery evidence, respectively. \\
\bottomrule
\end{tabularx}
\end{table}

\paragraph{Local execution.}
All roles use the \texttt{ContextualToolCallingAgent} wrapper around the smolagents \texttt{ToolCallingAgent}. They share the model $\mathcal{M}$ from Eq.~(\ref{eq:agent_node}) but receive different role instructions and action schemas. The history $\boldsymbol{h}_a^{(t)}$ represents the local interaction content recorded by \texttt{State} in the graph formulation. It begins with the invocation's instructions and input, including any explicitly supplied messages. For a nonterminal step, the model produces $\boldsymbol{y}_a^{(t)}$, the runtime executes its tool or managed-worker call, and the returned text is $\boldsymbol{x}_a^{\mathrm{obs},(t)}$. The history is updated as
\begin{equation}
\boldsymbol{h}_a^{(t+1)}=\boldsymbol{h}_a^{(t)}\mathbin{\Vert}\boldsymbol{y}_a^{(t)}\mathbin{\Vert}\boldsymbol{x}_a^{\mathrm{obs},(t)}.
\label{eq:appendix_local_history}
\end{equation}
This update preserves both the generated output and the returned observation. A tool result extends the invocation's local history; a worker report also carries spatial information from another invocation. The ordered history updates provide temporal information for later steps. A \texttt{final\_answer} action returns the invocation's result and terminates the run, so it does not require another tool observation. Textual observations here are distinct from the numerical $\boldsymbol{Q}/\boldsymbol{K}/\boldsymbol{V}$ observations used to compute $s_{a,\ell,r}$.

\paragraph{Delegation and communication.}
The manager has no direct Search, Visit, or Python tool. For open-Web tasks, the researcher and analyst are exposed as managed-agent actions; for closed-book tasks, only the analyst is exposed. A manager-to-worker call contains a natural-language subtask. The manager pauses while the selected worker executes its local ReAct loop and retains its own cache during this period. The worker's textual report becomes the observation of that delegation action, after which the manager resumes. Workers do not call one another. The returned object is serialized text---a final report and, for most backbones, a textual run summary---rather than hidden states or KV tensors. The worker's cache is released when its invocation finishes. The verifier is not callable by the manager: the runtime invokes it after the manager has returned its candidate answer and released its cache.

\paragraph{Agent-level runtime.}
Algorithm~\ref{alg:hierarchical_agent_runtime} describes the role-level workflow, while Algorithm~\ref{alg:aq_online_allocation} manages the caches of its invocations. The workflow includes at least one worker invocation and one verifier invocation. If verification requests more evidence or produces no usable answer, the runtime can launch one targeted recovery invocation, using the researcher for open-Web tasks or the analyst for closed-book tasks. The recovery evidence is then supplied with the manager's candidate to a fresh verifier invocation; this branch does not resume the completed manager. Model-specific parsing and answer-format repairs are omitted.

In the algorithm, \texttt{TaskClass} specifies the task-dependent configuration, \texttt{Team} stores the configured roles and permissions, and \texttt{Worker} identifies the selected recovery role. These are configuration records, not numerical quantities or the resident set $\mathcal{A}_r$. The notation $\boldsymbol{h}_{\mathrm{m}}$ is shorthand for the completed manager transcript, and $\boldsymbol{y}_{\mathrm{v}}$ denotes the latest verifier response, which need not yet be a final answer. All worker-result routines return serialized evidence, so evidence is combined by concatenation rather than set union. The configured model and permissions are used by every run; each new invocation has a separate identifier $a$.

\begin{algorithm}[t]
\caption{Agent-level execution used by the multi-agent runtime.}
\label{alg:hierarchical_agent_runtime}
\small
\begin{algorithmic}[1]
\Require Task query $\boldsymbol{q}$, \texttt{TaskClass}, model service $\mathcal{M}$
\Ensure Final prediction $\boldsymbol{z}$ in the requested format, or an explicit execution error
\State $\texttt{Team}\gets\Call{BuildTeam}{\texttt{TaskClass},\mathcal{M}}$
\State $(\widehat{\boldsymbol{y}},\boldsymbol{h}_{\mathrm{m}})\gets\Call{RunManager}{\boldsymbol{q},\texttt{Team}}$
\State $\boldsymbol{x}^{\mathrm{rep}}\gets\Call{WorkerReports}{\boldsymbol{h}_{\mathrm{m}}}$
\If{$\boldsymbol{h}_{\mathrm{m}}$ contains no managed-worker call}
    \State $\boldsymbol{x}^{\mathrm{rep}}\gets\Call{RunAnalystReview}{\boldsymbol{q},\widehat{\boldsymbol{y}}}$
\ElsIf{$\widehat{\boldsymbol{y}}$ explicitly reports missing evidence}
    \State $\texttt{Worker}\gets\Call{SelectRecoveryWorker}{\texttt{TaskClass}}$
    \State $\boldsymbol{x}^{\mathrm{rep}}\gets\boldsymbol{x}^{\mathrm{rep}}\mathbin{\Vert}\Call{RunWorker}{\texttt{Worker},\boldsymbol{q},\widehat{\boldsymbol{y}}}$
\EndIf
\State $\boldsymbol{y}_{\mathrm{v}}\gets\Call{RunVerifier}{\boldsymbol{q},\widehat{\boldsymbol{y}},\boldsymbol{x}^{\mathrm{rep}}}$
\If{$\boldsymbol{y}_{\mathrm{v}}$ requests more evidence or contains no usable answer}
    \State $\texttt{Worker}\gets\Call{SelectRecoveryWorker}{\texttt{TaskClass}}$
    \State $\boldsymbol{x}^{\mathrm{rec}}\gets\Call{RunTargetedRecovery}{\texttt{Worker},\boldsymbol{q},\widehat{\boldsymbol{y}},\boldsymbol{x}^{\mathrm{rep}},\boldsymbol{y}_{\mathrm{v}}}$
    \State $\boldsymbol{x}^{\mathrm{rep}}\gets\boldsymbol{x}^{\mathrm{rep}}\mathbin{\Vert}\boldsymbol{x}^{\mathrm{rec}}$
    \State $\boldsymbol{y}_{\mathrm{v}}\gets\Call{RunVerifier}{\boldsymbol{q},\widehat{\boldsymbol{y}},\boldsymbol{x}^{\mathrm{rep}}}$
\EndIf
\If{$\boldsymbol{y}_{\mathrm{v}}$ still contains no usable answer}
    \State $\boldsymbol{y}_{\mathrm{v}}\gets\Call{FinalVerifierAdjudication}{\boldsymbol{q},\widehat{\boldsymbol{y}},\boldsymbol{x}^{\mathrm{rep}},\boldsymbol{y}_{\mathrm{v}}}$
\EndIf
\State Accept the final answer from $\boldsymbol{y}_{\mathrm{v}}$ as $\boldsymbol{z}$, or raise an execution error
\State \Return $\boldsymbol{z}$
\end{algorithmic}
\end{algorithm}

\paragraph{From communication paths to cache branches.}
The implementation attaches a role-specific owner label of the form \texttt{role-r[rollout]-q[query-hash]} to each model request. This literal label identifies a role and rollout; it is not the allocation-round index $r$ or a replacement for the invocation identifier $a$. Different roles do not use one another's session identity. Entering a managed-agent call switches the active context from the manager to the worker; returning its report restores the manager context. Recovery calls use \texttt{reset=True}, so they start fresh local histories even when they reuse a role. Such a reset begins a new invocation rather than continuing the earlier invocation's private cache. In the full-history mode described here, each turn supplies the invocation's current transcript, and SGLang can reuse an unchanged prefix through its Radix cache. The optional role-owned streaming-session path is not used for these runs.

Communication across a spatial edge is serialized into task or report text. A worker report is tokenized again as part of the receiver's prompt; the worker's KV tensors are not copied to the receiver. The serving backend may also reuse identical physical prefixes across requests, but different role instructions, subtasks, reports, or tool observations produce separate suffixes. This backend-level reuse does not merge the invocation identities or transfer an agent's private reasoning state. The set $\mathcal{A}_r$ and Eq.~(\ref{eq:aq_reservation_main}) describe invocation-level residency and reservations, not a count of unique shared physical allocations. An invocation leaves the resident set when its private cache is released, while its returned text may remain in another invocation's history.

\begin{figure}[ht]
    \centering
    \includegraphics[width=\linewidth]{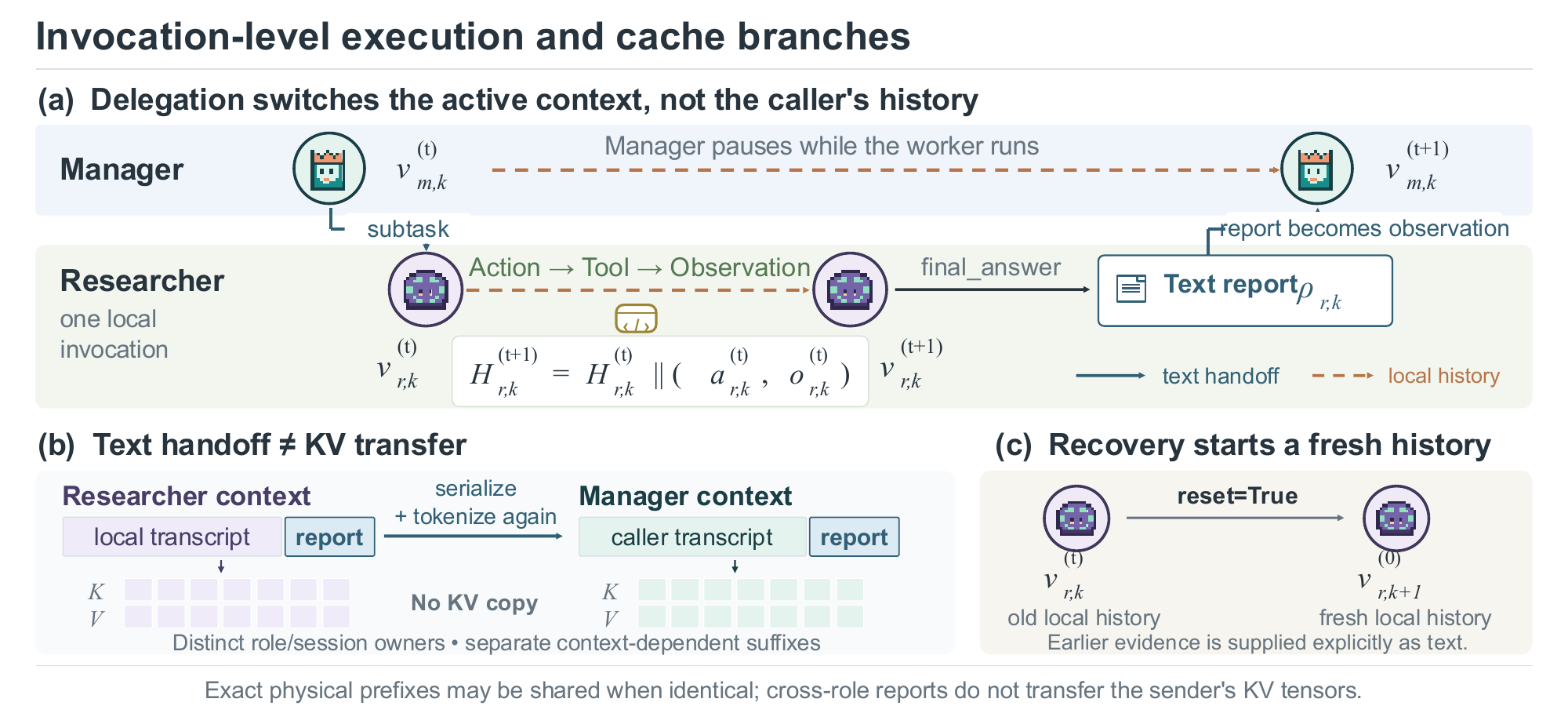}
    \caption{
    Invocation-level execution and cache branches.
    (a) The manager pauses during a managed-worker call while the worker extends its own local action--observation history; the returned textual report becomes the observation of the manager's delegation action.
    (b) Reports are serialized and tokenized as part of the receiver's prompt rather than transferred as KV tensors. Distinct context-dependent suffixes therefore form separate cache branches, although identical physical prefixes may be shared by the serving backend.
    (c) Recovery starts a fresh local history with \texttt{reset=True}; any earlier evidence is supplied explicitly as text.
    }
    \label{fig:runtime_cache_branches}
\end{figure}

Figure~\ref{fig:runtime_cache_branches} illustrates how local action--observation histories, textual handoffs, and fresh recovery invocations give rise to separate cache branches.

\section{Agent Roles and Prompt Construction}
\label{app:agent_prompts}

The task class determines which workers are exposed to the manager and which tools each role can call. Open-Web tasks use the manager, researcher, analyst, and verifier. Mathematical and closed-book tasks use the manager, analyst, and verifier; the researcher and all Web tools are removed. Table~\ref{tab:agent_tool_permissions} summarizes the actual action sets. The prompt clauses below use the portable JSON-action form described at the end of this section.

\begin{table}[htbp]
\centering
\captionsetup{justification=raggedright,singlelinecheck=false}
\caption{Role-specific actions in the multi-agent implementation. Every role also has \texttt{final\_answer}. The verifier's \texttt{VerificationSearchTool} is exposed under the name \texttt{search}, but it returns a fixed message and never accesses the network.}
\label{tab:agent_tool_permissions}
\small
\setlength{\tabcolsep}{3.5pt}
\renewcommand{\arraystretch}{1.12}
\begin{tabularx}{\linewidth}{@{}lXX@{}}
\toprule
\textbf{Role} & \textbf{Open-Web tasks} & \textbf{Closed-book tasks} \\
\midrule
Manager & Calls researcher or analyst & Calls analyst only \\
Researcher & Search, Visit, Python & Not active \\
Analyst & Python, Search, Visit & Python only \\
Verifier & Python and guarded Search & Python only \\
\bottomrule
\end{tabularx}
\end{table}

\begin{figure}[ht]
    \centering
    \includegraphics[width=\linewidth]{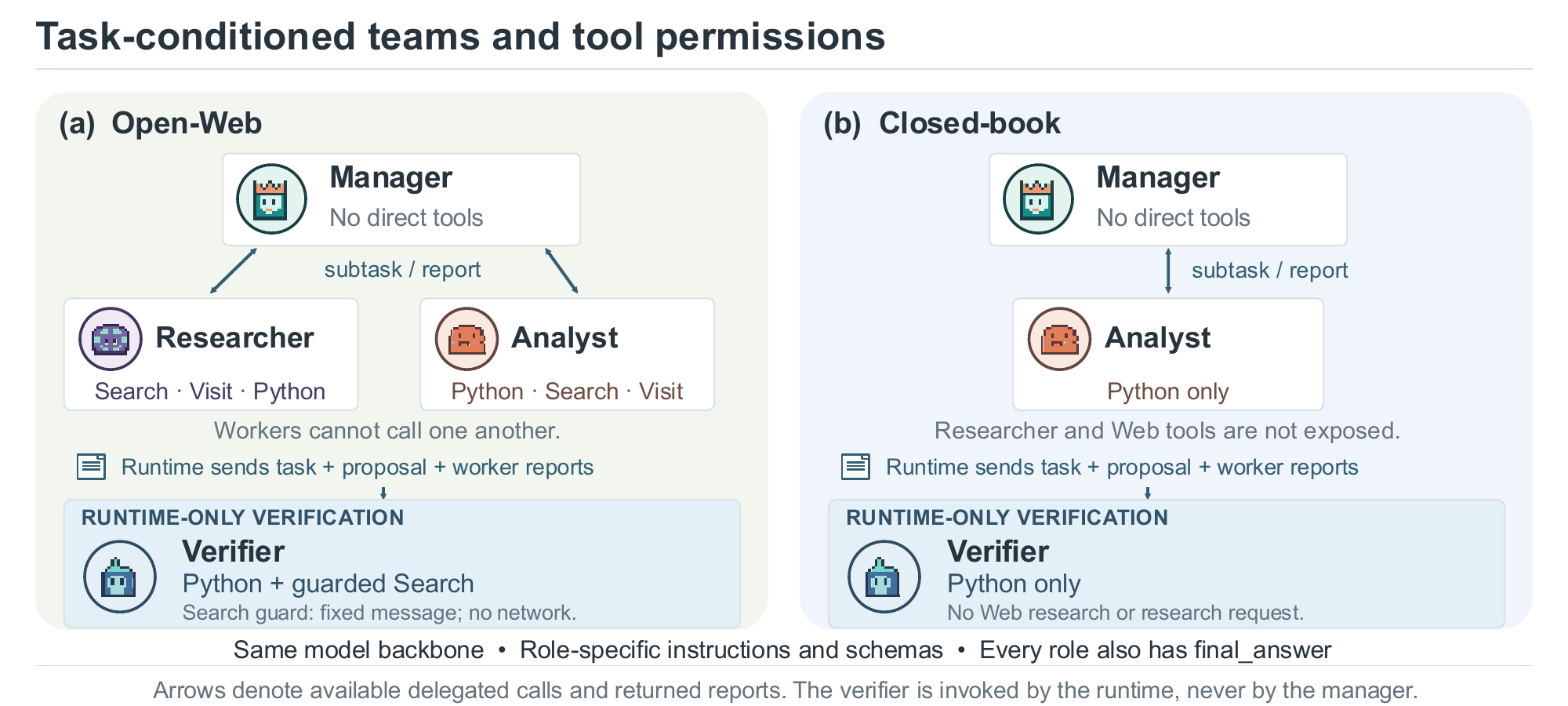}
    \caption{
    Task-conditioned teams and tool permissions.
    Open-Web tasks expose both researcher and analyst workers to the manager, whereas closed-book tasks expose only the analyst and disable Web tools.
    The verifier is a separate runtime stage, not a managed worker callable by the manager.
    Its guarded \texttt{search} action on open-Web tasks returns a fixed message without network access.
    Arrows show available task/report channels rather than a mandatory execution sequence, and every role also has \texttt{final\_answer}.
    }
    \label{fig:task_conditioned_teams}
\end{figure}

Figure~\ref{fig:task_conditioned_teams} visualizes the task-conditioned role and tool configuration in Table~\ref{tab:agent_tool_permissions}, distinguishing managed-\allowbreak worker calls from runtime-only verification.

\paragraph{Prompt assembly.}
For a fixed task class, the system prompt for role $i$ concatenates four text components:
\begin{equation}
\boldsymbol{p}_i^{\mathrm{sys}}=\boldsymbol{p}_{\mathrm{ReAct}}\mathbin{\Vert}\boldsymbol{p}_i^{\mathrm{schema}}\mathbin{\Vert}\boldsymbol{p}_i^{\mathrm{role}}\mathbin{\Vert}\boldsymbol{p}_i^{\mathrm{profile}}.
\label{eq:appendix_prompt_composition}
\end{equation}
Here, $\boldsymbol{p}_{\mathrm{ReAct}}$ explains the shared action--observation loop; $\boldsymbol{p}_i^{\mathrm{schema}}$ lists the permitted tools and managed-worker actions; $\boldsymbol{p}_i^{\mathrm{role}}$ describes the responsibility specified by $\texttt{Role}_i$; and $\boldsymbol{p}_i^{\mathrm{profile}}$ gives the task-specific instructions. The schema follows the available $\texttt{Tools}_i$ and the task-dependent worker permissions in Table~\ref{tab:agent_tool_permissions}. These symbols denote text components, not additional model functions; $\mathcal{P}$ retains its main-text meaning as the conditional output distribution. The task query or delegated subtask is supplied separately as the run input, and each invocation maintains its own $\boldsymbol{h}_a$. Roles therefore share $\mathcal{M}$ but differ in instructions, permissions, and local histories.

\begin{figure}[ht]
    \centering
    \includegraphics[width=\linewidth]{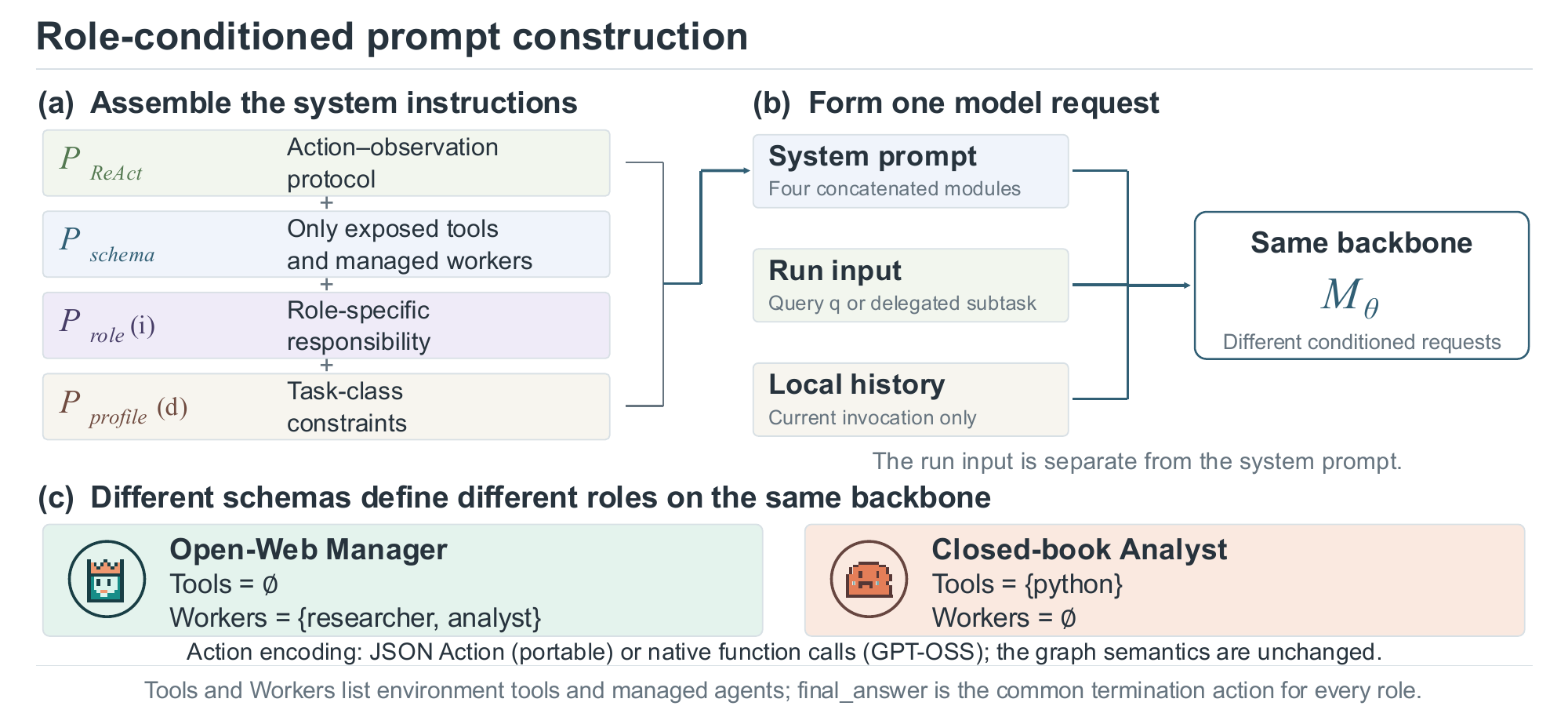}
    \caption{
    Role-conditioned prompt construction.
    The system prompt combines the shared ReAct protocol, the role's exposed action schemas, its responsibility, and the task-class profile.
    The query or delegated subtask is supplied separately as the run input, and each invocation retains its own local history.
    Roles use the same model backbone but differ in available actions, instructions, and visible histories.
    }
    \label{fig:prompt_assembly}
\end{figure}

Figure~\ref{fig:prompt_assembly} illustrates the composition in Eq.~(\ref{eq:appendix_prompt_composition}) and the separation between system instructions, run input, and local history.

\paragraph{Manager.}
The manager coordinates the team and produces a candidate answer. In multi-agent mode it has no direct environment tools. For open-Web tasks it may delegate to the researcher or analyst; for closed-book tasks it may delegate only to the analyst. The manager is instructed to use at least one worker, to combine returned reports, and to follow the requested answer format. Within the manager loop, the analyst can be called at most once and the researcher at most twice; a later runtime recovery is a separate invocation.

\begin{promptperiwinklebox}{Manager Role Clause for Open-Web Tasks}
Coordinate the team to solve the user task; do not solve it entirely by yourself. You must delegate each task to at least one appropriate specialist. Use researcher for web evidence and source-grounded facts, and analyst for calculations, parsing, closed-book reasoning, or consistency checks.

After gathering the relevant worker result(s), synthesize one proposed exact answer. The runtime will send that proposal to an independent verifier before it becomes the prediction. Call the analyst at most once and synthesize returned evidence instead of reopening the same reasoning loop. You may call the researcher a second time only when the first attempt hits an access block, lacks the requested exact fact, or exposes a source conflict.

Follow explicit output-format instructions in the original task literally. Keep reasoning concise and emit exactly one valid JSON Action per step before the token limit.
\end{promptperiwinklebox}

\paragraph{Researcher.}
The researcher receives a focused subtask, gathers evidence through search or page visits, and returns a textual report. Python is available for lightweight processing. Its report, not its private local history or KV states, is inserted into the manager's history.

\begin{promptperiwinklebox}{Researcher Role Clause}
Focus on evidence, URLs, dates, and exact factual claims. Do not invent sources. Prefer the primary source named in the task, including its archived revision, PDF, API, dataset, or repository. Never use benchmark pages, answer dumps, or model traces that merely quote a benchmark question or claimed answer as evidence; solve the task from the underlying source.

Keep reasoning concise and emit one valid JSON Action before the token limit; escape quotes inside JSON strings. Use one focused query per search call because searches share a global rate limit. You only own search, visit, python, and final_answer; never call researcher, analyst, or verifier. Do not repeat equivalent searches or revisit a blocked URL. By step 8, synthesize the best-supported answer from the evidence already collected unless one different primary-source attempt is likely to resolve the task. Stop searching once the evidence supports the answer and finish with final_answer.
\end{promptperiwinklebox}

\paragraph{Analyst.}
The analyst handles calculations, structured extraction, and consistency checks. On open-Web tasks it owns Python, Search, and Visit; on mathematical and closed-book tasks only Python is exposed. It cannot call another agent.

\begin{promptperiwinklebox}{Analyst Role Clause}
Prioritize calculations, parsing, and consistency checks. Show the relevant result and assumptions. When using Python, compute from the task inputs and inspect stdout, stderr, and returncode. Do not use Python merely to print an assumed candidate answer. Repair syntax or runtime errors before finishing.

Track units through every calculation. If the task asks for thousands, millions, percentages, or another scaled unit, convert the raw result before proposing the final value. Apply the requested rounding and separators exactly. Use only the tools exposed in your schema, never call another agent, and finish with final_answer.
\end{promptperiwinklebox}

\paragraph{Verifier.}
The verifier is a separate runtime stage, not one of the manager's managed agents. It receives the original task, manager proposal, and worker evidence. For open-Web tasks it cannot perform new Web research: its \texttt{search} action is a non-network guard. It may return a checked answer or identify a focused evidence gap with \texttt{NEEDS\_RESEARCH}; the runtime, rather than the verifier itself, launches any recovery worker. For closed-book tasks it independently reasons over the problem, may use Python, and does not request Web research. A new verifier invocation can receive additional evidence without reusing the preceding verifier's private cache.

\begin{promptperiwinklebox}{Verifier Role Clause for Open-Web Tasks}
Independently check the proposed answer against the supplied worker evidence and task constraints. Reject evidence that repeats a benchmark question or another agent's output. Require the primary source named by the task when it is accessible. Treat a bare search snippet, unsupported claim, or single source for a multi-hop or counting task as insufficient evidence. Request the missing hop, underlying rows, historical revision, or corroborating source.

Check arithmetic, physical units, requested scale, rounding, date and time formats, ordering, and separators. Do not invent a correction or start new research. If the evidence supports an answer, return only the corrected exact answer using final_answer. If a decisive fact is missing or sources conflict, return NEEDS_RESEARCH: <one focused evidence gap or query> using final_answer. Add no other explanation.
\end{promptperiwinklebox}

\begin{promptgreenbox}{Closed-Book Role Clauses}
Manager: Coordinate the solution with the analyst. Delegate every task to the analyst for independent reasoning, calculation, or Python verification. Web search, page visits, and the researcher are unavailable. Synthesize one exact candidate answer for the verifier.

Analyst: Reason independently and use Python when useful. Check calculations, units, signs, option mapping, and output format. Return the exact requested answer and concise supporting reasoning.

Verifier: Independently solve and check the manager proposal against the original task and analyst report. Use Python when useful, never request Web research, and return only the best-supported exact answer.
\end{promptgreenbox}

\paragraph{Action format.}
For the portable smolagents protocol, an agent emits one JSON Action naming either an owned tool, a managed worker, or \texttt{final\_answer}. A worker call creates a spatial edge, and a tool result extends the caller's temporal history.

\section{Task and Recovery Prompt Templates}
\label{app:runtime_prompts}

This section records the task-dependent inputs and textual handoffs used by Algorithm~\ref{alg:hierarchical_agent_runtime}. Table~\ref{tab:task_prompt_profiles} lists the task profiles for the four benchmarks named in Section~\ref{sec:experiments}. Generic examples from the underlying smolagents prompt are omitted.

\begin{table}[htbp]
\centering
\caption{Task profiles for the benchmark groups reported in the paper.}
\label{tab:task_prompt_profiles}
\small
\setlength{\tabcolsep}{4pt}
\renewcommand{\arraystretch}{1.12}
\begin{tabularx}{\linewidth}{@{}lX@{}}
\toprule
\textbf{Benchmarks} & \textbf{Agent organization and task instruction} \\
\midrule
Bamboogle, GAIA & Use the open-Web team. Delegate source discovery to the researcher and calculations or structured extraction to the analyst. \\
AIME25 & Use the closed-book team. Delegate derivation or Python verification to the analyst, disable Web access, and return only the requested exact numeric or symbolic expression. \\
GPQA & Use the closed-book team. Delegate scientific reasoning to the analyst, disable Web access, and return only the requested option, term, or short answer. \\
\bottomrule
\end{tabularx}
\end{table}

\paragraph{Messages passed between roles.}
Table~\ref{tab:agent_message_payloads} lists the textual payloads and their delivery routes. Runtime-mediated routes describe how messages are delivered, not additional agent vertices in $\mathcal{G}_{\boldsymbol{q}}$. The payloads become inputs to the receiving invocation, consistent with $\boldsymbol{x}_i^{\mathrm{S},(t)}$ in Eq.~(\ref{eq:agent_communication}). KV tensors are not transferred; a sender's private cache is released when its invocation ends.

\begin{table}[htbp]
\centering
\caption{Textual payloads passed along the agent graph.}
\label{tab:agent_message_payloads}
\small
\setlength{\tabcolsep}{4.5pt}
\renewcommand{\arraystretch}{1.12}
\begin{tabularx}{\linewidth}{@{}lX@{}}
\toprule
\textbf{Message route} & \textbf{Payload visible to the receiver} \\
\midrule
Manager $\rightarrow$ worker & A delegated subtask and optional additional arguments. \\
Worker $\rightarrow$ manager & The worker's final textual report, optionally followed by a textual run summary. \\
Runtime $\rightarrow$ verifier & Original task, manager proposal, and extracted worker report(s). \\
Verifier $\rightarrow$ recovery worker (via runtime) & One focused evidence gap, together with the original task and earlier candidate/evidence. \\
Recovery worker $\rightarrow$ verifier (via runtime) & Recovery summary and the relevant Search, Visit, or Python observations. \\
\bottomrule
\end{tabularx}
\end{table}

\begin{figure}[ht]
    \centering
    \includegraphics[width=\linewidth]{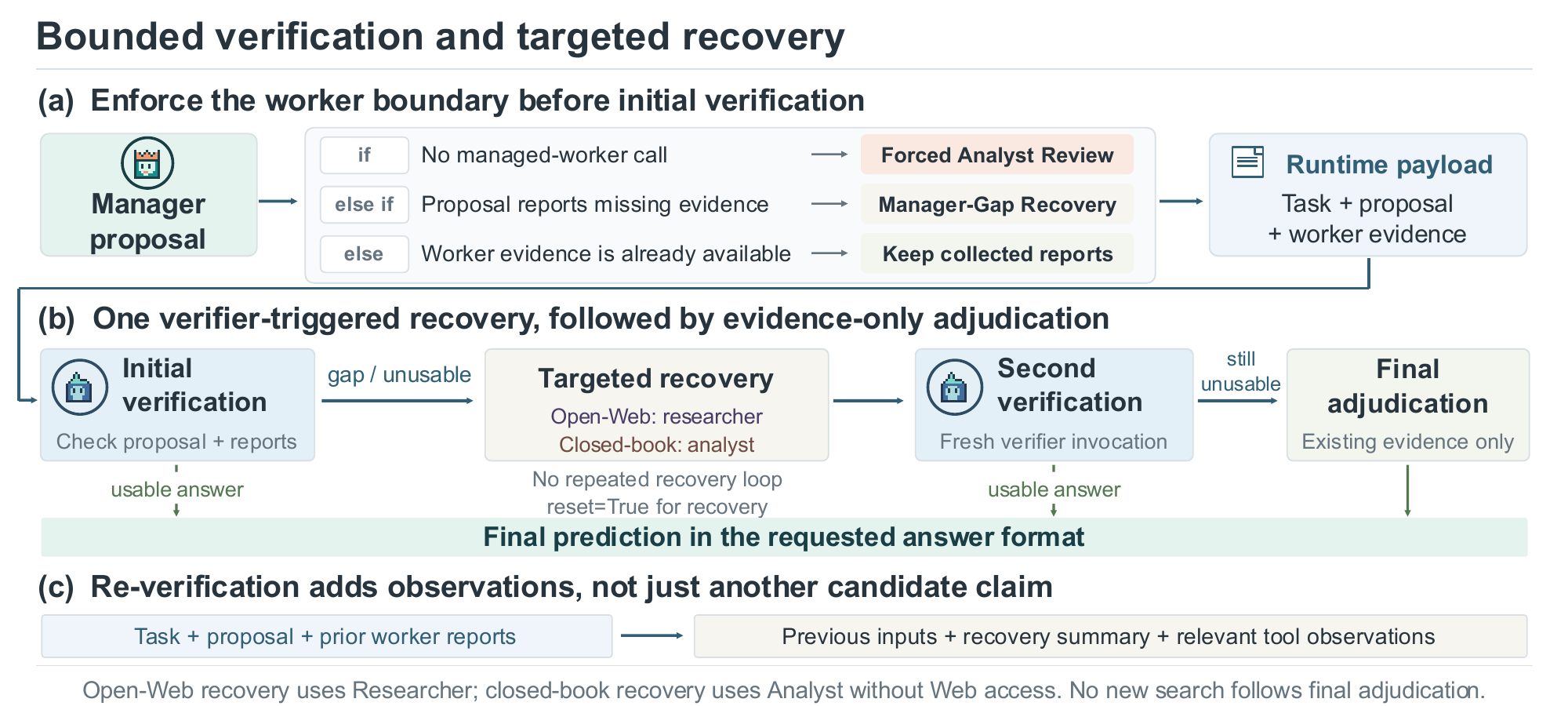}
    \caption{
    Bounded verification and targeted recovery.
    (a) Before initial verification, the runtime performs a forced analyst review if no worker was called; otherwise, an explicitly uncertain manager proposal can trigger manager-gap recovery.
    (b) An evidence request or unusable verification output triggers at most one targeted recovery in this branch, using a researcher for open-Web tasks or an analyst without Web access for closed-book tasks. A new verifier invocation checks the augmented evidence, with a final evidence-only adjudication if necessary and no further search branch.
    (c) Re-verification receives relevant tool observations together with the recovery summary, rather than treating repeated candidate claims as evidence.
    }
    \label{fig:verification_recovery}
\end{figure}

Figure~\ref{fig:verification_recovery} summarizes the bounded verification and recovery workflow in Algorithm~\ref{alg:hierarchical_agent_runtime}, providing an overview of when the following prompt templates are used.

\paragraph{Managed-worker call.}
The manager sends a worker a \texttt{task} and optional \texttt{additional\_args} through the managed-agent schema. The library adds the worker's name and asks it to return a report for the manager. In simplified form, the handoff is:

\begin{promptperiwinklebox}{Managed-Agent Handoff}
You are the [RESEARCHER or ANALYST].

Task delegated by the manager:
[SUBTASK]

Use only the tools in your schema. Return a textual report that the manager can use.
\end{promptperiwinklebox}

\paragraph{Enforcing the worker boundary.}
If the manager returns a candidate without calling any worker, the runtime starts a fresh analyst invocation before verification. Its returned review forms $\boldsymbol{x}^{\mathrm{rep}}$ and is supplied to the verifier together with $\widehat{\boldsymbol{y}}$; the completed manager is not resumed:

\begin{prompttealbox}{Forced Analyst Review after Manager Bypass}
Review the proposed answer for correctness, calculations, formatting, and compliance with the original task. Return a corrected exact answer if needed.

Original task:
[QUESTION]

Manager proposal:
[MANAGER PROPOSAL]
\end{prompttealbox}

\paragraph{Recovery from an uncertain manager proposal.}
If the manager has already used a worker but its candidate explicitly reports missing evidence, the runtime invokes one worker again before verification: the researcher for open-Web tasks or the analyst for closed-book tasks. The returned text is appended to $\boldsymbol{x}^{\mathrm{rep}}$ and passed to the verifier with the existing candidate.

\begin{prompttealbox}{Manager-Gap Recovery Input}
Revisit the uncertain manager proposal once.

For an open-Web task, use a different focused query or an alternative accessible source and return the exact answer with the shortest supporting evidence.

For a closed-book task, perform an independent reasoning and Python verification pass without Web access and return the exact best-supported answer.

Original task:
[QUESTION]

Uncertain manager proposal:
[MANAGER PROPOSAL]
\end{prompttealbox}

\paragraph{Initial verification.}
The runtime starts verification after the manager and any required review have finished. The verifier receives the original task $\boldsymbol{q}$, candidate $\widehat{\boldsymbol{y}}$, and collected evidence $\boldsymbol{x}^{\mathrm{rep}}$, but no other invocation's KV tensors. Its response is denoted by $\boldsymbol{y}_{\mathrm{v}}$ until the runtime accepts a final prediction $\boldsymbol{z}$.

\begin{promptperiwinklebox}{Open-Web Initial Verification Input}
Independently verify the proposed answer against the original task and worker evidence. Return only the corrected exact answer when the evidence directly supports it. If one decisive fact is missing or conflicting, return NEEDS_RESEARCH: <one focused evidence gap or query>. Do not guess or include any other explanation.

Original task:
[QUESTION]

Manager proposal:
[MANAGER PROPOSAL]

Worker review:
[WORKER EVIDENCE]
\end{promptperiwinklebox}

\begin{promptgreenbox}{Closed-Book Initial Verification Input}
Independently solve and verify this closed-book problem against the Manager proposal and Analyst review. Use Python when useful, choose the best-supported option or value, and return only the exact answer requested. Do not search, request research, abstain, or include an explanation.

Original task:
[QUESTION]

Manager proposal:
[MANAGER PROPOSAL]

Analyst review:
[WORKER EVIDENCE]
\end{promptgreenbox}

\paragraph{Targeted recovery.}
For open-Web tasks, an evidence request or unusable verifier response can trigger a fresh researcher invocation. For closed-book tasks, an unusable response triggers a fresh analyst invocation without Web access. The runtime supplies the original task, candidate, earlier evidence, and verifier response to the selected worker. The returned recovery summary and relevant tool observations form $\boldsymbol{x}^{\mathrm{rec}}$.

\begin{prompttealbox}{Targeted Recovery Input}
Resolve the verifier's specific evidence gap.

For an open-Web task, use a different focused query or source when earlier evidence is blocked or inconclusive, and inspect the named primary page, PDF, historical revision, dataset, or repository when possible.

For a closed-book task, perform an independent derivation and Python check without Web access.

Return the exact answer plus the shortest decisive quotation or calculation. Do not guess or abstain.

Original task:
[QUESTION]

Manager proposal:
[MANAGER PROPOSAL]

Earlier worker review:
[WORKER EVIDENCE]

Verifier evidence gap:
[VERIFIER OUTPUT]
\end{prompttealbox}

\paragraph{Verification after recovery.}
The runtime appends $\boldsymbol{x}^{\mathrm{rec}}$ to the previously collected evidence and supplies the result, together with $\boldsymbol{q}$ and $\widehat{\boldsymbol{y}}$, to a fresh verifier invocation. The recovery report is not sent back to the completed manager. Relevant tool observations accompany the recovery summary so that the verifier can check its claims against the supplied evidence. The following template shows the open-Web case. Closed-book verification retains the no-Web instruction given above and uses the additional analyst evidence.

\begin{prompttealbox}{Open-Web Verification after Recovery}
Re-check the original task after targeted evidence recovery. Return only the exact final answer when supported. Treat the manager proposal and recovery summary as candidate claims. Validate them against the actual Search, Visit, or Python observations. If the decisive fact remains unavailable, return NEEDS_RESEARCH: <remaining evidence gap>.

Original task:
[QUESTION]

Manager proposal:
[MANAGER PROPOSAL]

Earlier worker evidence:
[WORKER EVIDENCE]

Targeted recovery evidence:
[RECOVERY EVIDENCE]
\end{prompttealbox}

If the second verification is still unusable, the runtime invokes the verifier role once more for evidence-only adjudication over the collected messages; this does not start another search branch. The accepted answer becomes $\boldsymbol{z}$; an unresolved execution failure is not treated as a valid prediction. Bounded parser, terminal-error, and answer-format retries are implementation safeguards rather than additional agent roles, so their prompt text is not reproduced here.

\section{Derivations and Algorithm Details}
\label{app:aq_details}

This appendix gives the assumptions behind the attention bound, proves the allocation statements, and specifies the storage contract. No calibration dataset is used. The numerical observations come from ordinary forwards in the current invocation. The theoretical bound concerns ideal-grid quantization of an observed attention window; using its FP32 coefficient to choose precision for future blocks is a planning approximation.

\subsection{From Attention to the Estimated Quantization Error}
\label{app:aq_attention_to_objective}

This subsection connects the attention computation in Eq.~(\ref{eq:aq_attention}) to the allocation objective in Eq.~(\ref{eq:aq_estimated_error}). We first bound the effect of rounding keys and values within an observed window, then square and average this bound within each layer to obtain $s_{a,\ell,r}$. Finally, we explain the assumptions used to apply the local score to prospective blocks.

\paragraph{1. Local attention and assumptions.}
Fix invocation $a$ and the observation used to compute its score at allocation round $r$. For one query and attention head, use $\boldsymbol{q}\in\mathbb R^d$ and $\boldsymbol{K},\boldsymbol{V}\in\mathbb R^{w\times d}$ as in the main text. Let $\widetilde{\boldsymbol{K}}_b$ and $\widetilde{\boldsymbol{V}}_b$ be the reconstructed $b$-bit matrices, where $b\ge1$ is an integer. The query and visible positions remain unchanged. Define the query--key score vector $\boldsymbol{z}=\boldsymbol{K}\boldsymbol{q}/\sqrt d$ and attention weights $\boldsymbol{p}=\operatorname{Softmax}(\boldsymbol{z})$. Their quantized counterparts are $\widetilde{\boldsymbol{z}}_b=\widetilde{\boldsymbol{K}}_b\boldsymbol{q}/\sqrt d$ and $\widetilde{\boldsymbol{p}}_b=\operatorname{Softmax}(\widetilde{\boldsymbol{z}}_b)$. Thus,
\begin{equation}
\boldsymbol{o}=\boldsymbol{p}^{\top}\boldsymbol{V},\qquad
\widetilde{\boldsymbol{o}}_b=\widetilde{\boldsymbol{p}}_b^{\top}\widetilde{\boldsymbol{V}}_b,
\qquad \boldsymbol{o},\widetilde{\boldsymbol{o}}_b\in\mathbb R^{1\times d}.
\label{eq:aq_app_attention_outputs}
\end{equation}
We assume real arithmetic, exact group endpoints, and nearest rounding to $2^b$ uniformly spaced levels. The attention weights are normalized over the observed visible positions, which must form a nonempty set. For a fixed mask, the proof applies to the visible positions while retaining the original quantization groups and their ranges.

\paragraph{Proposition 1 (window-local attention bound).}
Under these assumptions, Eq.~(\ref{eq:aq_attention_bound}) holds with $\kappa,\nu,\rho$ defined in Eq.~(\ref{eq:aq_kappa_nu}). Steps 2--4 below give the proof.

\paragraph{2. Bound the errors in keys, values, and query--key scores.}
A $b$-bit quantizer divides a group's range into $2^b-1$ equal intervals. Nearest rounding therefore changes an entry by at most half an interval. Using the same groups as in the main text gives
\begin{equation}
\begin{aligned}
\bigl|[\widetilde{\boldsymbol{K}}_b-\boldsymbol{K}]_{ij}\bigr|
&\le\frac{\max_{i'\in\mathcal{G}}[\boldsymbol{K}]_{i'j}-\min_{i'\in\mathcal{G}}[\boldsymbol{K}]_{i'j}}{2(2^b-1)},\\
\bigl|[\widetilde{\boldsymbol{V}}_b-\boldsymbol{V}]_{ij}\bigr|
&\le\frac{\max_{j'\in\mathcal{G}}[\boldsymbol{V}]_{ij'}-\min_{j'\in\mathcal{G}}[\boldsymbol{V}]_{ij'}}{2(2^b-1)}.
\end{aligned}
\label{eq:aq_app_grid}
\end{equation}
In the first line, $\mathcal{G}$ is the key-token group containing $i$; in the second, it is the value-feature group containing $j$. The primed indices run over entries within that group. No separate variable is needed for either group range.

Since the query is fixed, each query--key score changes only through key rounding. Applying the triangle inequality and then the definition of $\kappa$ gives
\begin{equation}
\begin{aligned}
\|\widetilde{\boldsymbol{z}}_b-\boldsymbol{z}\|_{\infty}
&\le\frac{1}{\sqrt d}\max_i\sum_j|q_j|\bigl|[\widetilde{\boldsymbol{K}}_b-\boldsymbol{K}]_{ij}\bigr|\\
&\le\frac{\sum_j|q_j|\max_{\mathcal{G}}\bigl(\max_{i'\in\mathcal{G}}[\boldsymbol{K}]_{i'j}-\min_{i'\in\mathcal{G}}[\boldsymbol{K}]_{i'j}\bigr)}{2(2^b-1)\sqrt d}
=\frac{\kappa}{2(2^b-1)}.
\end{aligned}
\label{eq:aq_app_key_scores}
\end{equation}
For values, every feature in a group has the same rounding-error bound. Summing squared entry bounds therefore contributes the group's actual size $|\mathcal{G}|$. With $[\boldsymbol{V}]_{i,:}$ denoting row $i$, we obtain
\begin{equation}
\begin{aligned}
\max_i\bigl\|[\widetilde{\boldsymbol{V}}_b-\boldsymbol{V}]_{i,:}\bigr\|_2
&\le\frac{\max_i\sqrt{\sum_{\mathcal{G}}|\mathcal{G}|\bigl(\max_{j\in\mathcal{G}}[\boldsymbol{V}]_{ij}-\min_{j\in\mathcal{G}}[\boldsymbol{V}]_{ij}\bigr)^2}}{2(2^b-1)}\\
&=\frac{\nu}{2(2^b-1)}.
\end{aligned}
\label{eq:aq_app_value_rounding}
\end{equation}
These bounds use the complete observed groups. Restricting attention to fewer visible rows cannot increase the resulting maxima.

\paragraph{3. Bound the change in attention weights.}
To pass from query--key scores to attention weights, consider the softmax Jacobian $\boldsymbol{J}_{\boldsymbol{p}}=\operatorname{diag}(\boldsymbol{p})-\boldsymbol{p}\boldsymbol{p}^{\top}$. For any perturbation vector $\boldsymbol{\delta}$, its $i$-th output entry is $p_i(\delta_i-\boldsymbol{p}^{\top}\boldsymbol{\delta})$. Since $p_i\ge0$ and $\sum_i p_i=1$,
\begin{equation}
\begin{aligned}
\|\boldsymbol{J}_{\boldsymbol{p}}\boldsymbol{\delta}\|_1
&=\sum_i p_i\bigl|\delta_i-\boldsymbol{p}^{\top}\boldsymbol{\delta}\bigr|\\
&\le\sqrt{\sum_i p_i\bigl(\delta_i-\boldsymbol{p}^{\top}\boldsymbol{\delta}\bigr)^2}\\
&=\sqrt{\sum_i p_i\delta_i^2-(\boldsymbol{p}^{\top}\boldsymbol{\delta})^2}
\le\|\boldsymbol{\delta}\|_{\infty}.
\end{aligned}
\label{eq:aq_app_softmax}
\end{equation}
The first inequality is the Cauchy--Schwarz inequality with weights $p_i$. The last inequality follows from $\sum_i p_i\delta_i^2\le\|\boldsymbol{\delta}\|_{\infty}^2$. Applying this Jacobian bound along the line segment from $\boldsymbol{z}$ to $\widetilde{\boldsymbol{z}}_b$ gives
\begin{equation}
\|\widetilde{\boldsymbol{p}}_b-\boldsymbol{p}\|_1
\le\|\widetilde{\boldsymbol{z}}_b-\boldsymbol{z}\|_{\infty}
\le\frac{\kappa}{2(2^b-1)}.
\label{eq:aq_app_attention_weights}
\end{equation}
Thus, $\kappa$ controls the possible change in attention weights caused by key rounding.

\paragraph{4. Combine the two effects on the attention output.}
Adding and subtracting $\widetilde{\boldsymbol{p}}_b^{\top}\boldsymbol{V}$ separates the output error into two terms:
\begin{equation}
\widetilde{\boldsymbol{o}}_b-\boldsymbol{o}
=(\widetilde{\boldsymbol{p}}_b-\boldsymbol{p})^{\top}\boldsymbol{V}
+\widetilde{\boldsymbol{p}}_b^{\top}(\widetilde{\boldsymbol{V}}_b-\boldsymbol{V}).
\label{eq:aq_error_split}
\end{equation}
The first term changes the weights while holding values fixed; the second changes the values under the quantized weights. To bound the first term, let $\boldsymbol{c}\in\mathbb R^{1\times d}$ contain the coordinate midpoints of the value vectors:
\begin{equation}
c_j=\frac{\max_i[\boldsymbol{V}]_{ij}+\min_i[\boldsymbol{V}]_{ij}}{2},
\qquad
\max_i\|[\boldsymbol{V}]_{i,:}-\boldsymbol{c}\|_2
\le\frac{1}{2}\sqrt{\sum_j\bigl(\max_i[\boldsymbol{V}]_{ij}-\min_i[\boldsymbol{V}]_{ij}\bigr)^2}
=\rho.
\label{eq:aq_app_value_center}
\end{equation}
Both attention-weight vectors sum to one, so $\sum_i(\widetilde p_{b,i}-p_i)=0$. Subtracting the same midpoint from every value row therefore leaves the first term unchanged:
\begin{equation}
\begin{aligned}
\|(\widetilde{\boldsymbol{p}}_b-\boldsymbol{p})^{\top}\boldsymbol{V}\|_2
&=\left\|\sum_i(\widetilde p_{b,i}-p_i)([\boldsymbol{V}]_{i,:}-\boldsymbol{c})\right\|_2\\
&\le\sum_i|\widetilde p_{b,i}-p_i|\,\|[\boldsymbol{V}]_{i,:}-\boldsymbol{c}\|_2
\le\rho\|\widetilde{\boldsymbol{p}}_b-\boldsymbol{p}\|_1.
\end{aligned}
\label{eq:aq_app_weight_effect}
\end{equation}
For the second term, the quantized attention weights remain nonnegative and sum to one. Its norm is therefore at most the largest value-row rounding error. Combining both terms yields
\begin{equation}
\begin{aligned}
\|\widetilde{\boldsymbol{o}}_b-\boldsymbol{o}\|_2
&\le\rho\|\widetilde{\boldsymbol{p}}_b-\boldsymbol{p}\|_1
+\sum_i\widetilde p_{b,i}\bigl\|[\widetilde{\boldsymbol{V}}_b-\boldsymbol{V}]_{i,:}\bigr\|_2\\
&\le\rho\|\widetilde{\boldsymbol{p}}_b-\boldsymbol{p}\|_1
+\max_i\bigl\|[\widetilde{\boldsymbol{V}}_b-\boldsymbol{V}]_{i,:}\bigr\|_2\\
&\le\frac{\rho\kappa+\nu}{2(2^b-1)}.
\end{aligned}
\label{eq:aq_app_output_bound}
\end{equation}
This proves Eq.~(\ref{eq:aq_attention_bound}) without assuming independent or zero-mean quantization errors. The factor $\rho$ is needed because changes in attention weights matter only through differences between the value vectors. If all value rows are identical, $\rho=0$ and the first error term vanishes.

\paragraph{5. Square and average to obtain the sensitivity score.}
Squaring the nonnegative bound gives
\begin{equation}
\|\widetilde{\boldsymbol{o}}_b-\boldsymbol{o}\|_2^2
\le\frac{(\rho\kappa+\nu)^2}{4(2^b-1)^2}.
\label{eq:aq_app_squared_bound}
\end{equation}
The coefficient $(\rho\kappa+\nu)^2/4$ depends on the observed computation, whereas $(2^b-1)^{-2}$ depends only on precision. We compute the coefficient for each observed layer $\ell$, query head $h$, and query position $u$, and average within the layer as in Eq.~(\ref{eq:aq_online_score}). Let $\mathcal{H}(h)$ identify the KV head used by query head $h$. Then $\kappa_{\ell,h,u}$ uses that query and its associated key head, while $\rho_{\ell,\mathcal{H}(h)}$ and $\nu_{\ell,\mathcal{H}(h)}$ use the corresponding value head. The layer score is
\begin{equation}
s_{a,\ell,r}
=\frac{\displaystyle\sum_{h,u}\bigl(\rho_{\ell,\mathcal{H}(h)}\kappa_{\ell,h,u}+\nu_{\ell,\mathcal{H}(h)}\bigr)^2}{\displaystyle4\sum_{h,u}1}.
\label{eq:aq_score_expanded}
\end{equation}
The sums enumerate the observed head--query pairs in layer $\ell$; the factor $1/4$ comes from the squared bound. Query heads sharing one KV head reuse its key and value statistics but retain their own query-dependent $\kappa$. For one layer and candidate width $b$, the bit-dependent factor can be taken outside the average:
\begin{equation}
\operatorname{Average}_{h,u}\!\left[\|\widetilde{\boldsymbol{o}}_{\ell,h,u,b}-\boldsymbol{o}_{\ell,h,u}\|_2^2\right]
\le\frac{s_{a,\ell,r}}{(2^b-1)^2}.
\label{eq:aq_app_average_error}
\end{equation}
Squaring precedes averaging. This bounds the mean of the squared local output errors, not the squared error after combining all heads or propagating through later layers.

\paragraph{6. Construct the estimate over remaining blocks and invocations.}
The inequality above concerns the recorded observation window. To use it for allocation, we treat $s_{a,\ell,r}$ as a surrogate for prospective blocks in that layer and evaluate $s_{a,\ell,r}/(2^{b_{a,\ell}}-1)^2$. This extrapolates from current observations; a future block need not have the same entries or quantization groups.

Eq.~(\ref{eq:aq_remaining_blocks}) gives remaining quantizable capacity per layer; it is used in the byte reservation, not to predict generation length. The audited controller ranks layer scores and minimizes the additive surrogate
\begin{equation}
\widehat{\mathcal{E}}_r(\boldsymbol{b})
=\sum_{a\in\mathcal{A}_r}\sum_{\ell\in\mathcal{L}}
\frac{s_{a,\ell,r}}{(2^{b_{a,\ell}}-1)^2}.
\label{eq:aq_app_total_estimate}
\end{equation}
This is an allocation objective, not a bound on final-answer error. Already sealed blocks are excluded from its adjustable component because their precision cannot change; their bytes still count toward the reservation. A layer with no current observation receives the median valid resident score, or $1$ when no score exists.

The bound applies to attention computed within the observed window under exact group endpoints and nearest rounding. Assuming that recent observations remain representative of the remaining blocks, we use this bound to construct the additive allocation objective in Eq.~(\ref{eq:aq_estimated_error}), rather than a bound on final-answer error.

\subsection{Memory Reservation Calculation}
\label{app:aq_storage_details}

We detail the four terms in Eq.~(\ref{eq:aq_reservation_main}). All costs are measured in bytes, and the $L$ quantized layers are assumed to share the same cache geometry. The formula reserves memory for existing storage and further cache growth, rather than measuring only the current occupancy.

\paragraph{Storage of one quantized block.}
Let $H$ and $H_{\mathrm{q}}$ denote the numbers of KV heads and query heads. Each block contains $P$ tokens per layer, with $d_{\mathrm{K}}$ key features and $d_{\mathrm{V}}$ value features per KV head. With group size $G$, keys are grouped across tokens within each feature, and values across features within each token. A block stored at $b$ bits occupies
\begin{equation}
\begin{aligned}
\mathcal{B}_{\mathrm{blk}}(b)
&=H\left(\left\lceil\frac{Pd_{\mathrm{K}}b}{8}\right\rceil+\left\lceil\frac{Pd_{\mathrm{V}}b}{8}\right\rceil\right)\\
&\quad+4H\left(\left\lceil\frac{P}{G}\right\rceil d_{\mathrm{K}}+P\left\lceil\frac{d_{\mathrm{V}}}{G}\right\rceil\right).
\end{aligned}
\label{eq:aq_app_block_bytes}
\end{equation}
The first line counts packed key and value codes, with ceilings accounting for the final partial byte in each head's payload. The implementation divides each code across a fixed base payload and zero or more one-bit refinement planes. The audited 2-bit target uses a physical 1-bit base, so a 1-bit block consumes fewer packed-code bytes than a 2-bit block and can earn byte credit for a 3-bit block. The 3- and 4-bit targets use 2- and 3-bit bases. Refinement storage is committed on demand; the reservation formula counts the selected code width, while the physical audit also reports committed refinement storage. The second line counts $H\lceil P/G\rceil d_{\mathrm{K}}$ key groups and $HP\lceil d_{\mathrm{V}}/G\rceil$ value groups. Each group stores a two-byte scale and a two-byte offset, giving four bytes per group.

\paragraph{Existing and additional quantized storage.}
The term $I_{a,r}$ sums codes, scales, and offsets in already quantized blocks across layers at their stored widths. Given capacity $T_a$, a native-precision recent window of $R$ tokens, and $z_{a,\ell,r}$ already quantized tokens in layer $\ell$, Eq.~(\ref{eq:aq_remaining_blocks}) gives its remaining block count $n_{a,\ell,r}$. The additional quantized reservation is $\sum_{\ell}n_{a,\ell,r}\mathcal{B}_{\mathrm{blk}}(b_{a,\ell})$.

\paragraph{Unquantized states and observations.}
Let $u$ be the native bytes per scalar, $A$ the maximum tokens appended in one forward pass, $W$ the observation-window limit, and $\tau_{a,r}$ the current unquantized tail length. We reserve
\begin{equation}
\begin{aligned}
U_{a,r}&=LuH(d_{\mathrm{K}}+d_{\mathrm{V}})\min\!\left\{T_a-z_{a,r},\max\!\left(\tau_{a,r},P-1+R+A\right)\right\},\\
O_a&=Lu\left[H_{\mathrm{q}}d_{\mathrm{K}}\min(W,A,T_a)+H(d_{\mathrm{K}}+d_{\mathrm{V}})\min(W,T_a)\right].
\end{aligned}
\label{eq:aq_app_raw_observation}
\end{equation}
In $U_{a,r}$, $R+P-1+A$ allows for recent tokens, an incomplete block, and one forward pass's new entries. The maximum also covers the current tail, while the minimum limits the allowance to capacity not already quantized. Multiplying by $LuH(d_{\mathrm{K}}+d_{\mathrm{V}})$ converts the token allowance to bytes. In $O_a$, the first term covers current query observations and the second covers recent key and value observations. This space remains reserved between observations.

\paragraph{Total reservation.}
Adding existing quantized storage, additional quantized capacity, unquantized states, and observations gives Eq.~(\ref{eq:aq_reservation_main}). The additional packed capacity and unquantized allowance intentionally overlap because entries are stored at native precision before quantization; the reservation therefore includes both forms. Observation tensors are counted once. The strict average-width envelope in Eq.~(\ref{eq:aq_target_envelope}) compares this packed-KV reservation with the same reservation evaluated at the target width for all sealed and prospective blocks. It is a KV-storage budget, not a bound on total process GPU memory. Model weights, temporary attention and codec workspace, Python records, and allocator overhead are excluded.

\subsection{Balanced Agent--Layer Allocation}
\label{app:aq_allocation_details}

For equal-sized KV layers, a lower-bit and a higher-bit choice have opposite, equal packed-code byte deviations from the target. We pair these choices \emph{within each invocation}, so the exact target packed-code mean does not depend on the lifetime of another agent. Let $s_{a,(1),r}\le\cdots\le s_{a,(L),r}$ be the layer sensitivities in ascending order. At a fixed pair count $k$, the optimal arrangement for Eq.~(\ref{eq:aq_estimated_error}) assigns $b_{\mathrm{avg}}-1$ to the $k$ smallest scores and $b_{\mathrm{avg}}+1$ to the $k$ largest scores. This follows by exchanging any inverted pair: the bit-dependent error coefficient decreases as the bit width increases, so assigning the larger width to the larger score cannot increase the objective. The remaining $L-2k$ layers use $b_{\mathrm{avg}}$. We evaluate $k=1,\ldots,\lfloor(L-1)/2\rfloor$ and choose the lowest estimated error, breaking ties toward fewer exchanges. The lower bound guarantees all three candidate widths, and the upper bound keeps at least one target-width layer. Scores tied before any observation are ordered by a deterministic invocation--layer hash. If a new vector duplicates one already assigned or physically sealed by a teammate, we choose the distinct vector reachable by one swap of differently coded layers with the smallest increase in the surrogate. This tie repair preserves the width counts and byte budget.

Historical sealed blocks retain their widths. The reservation in Eq.~(\ref{eq:aq_reservation_main}) includes those historical bytes, future blocks at the selected layer widths, the raw tail, and observations. The controller compares the reservation with the configured team budget and the average-width entitlement at every allocation event. The physical audit separately checks all sealed widths, the exact mean packed-code width, and compressed KV storage; an allocation proposal by itself does not certify a valid result. This constrained selection is the audited end-to-end policy. The earlier agent-level Pareto solver remains available in code for controls and ablations but is not used to generate the audited agent--layer result cells.

\section{End-to-end Algorithm}
\label{app:detailed_algorithm}

Algorithm~\ref{alg:aq_online_allocation} presents the end-to-end procedure of \method{}, combining online sensitivity estimation, joint bit-width allocation under a shared memory budget, and quantization of newly eligible blocks. It also describes cache initialization, prefix reuse, and cache release as agent invocations progress. Updated bit widths apply only to newly quantized blocks, while retained quantized blocks remain unchanged.

\begin{algorithm}[htbp]
\caption{End-to-end agent--layer KV-cache quantization in \method{}}
\label{alg:aq_online_allocation}
\small
\begin{algorithmic}[1]
\Require Multi-agent runtime, resident set $\mathcal{A}$, target $b_{\mathrm{avg}}$, layer set $\mathcal{L}$
\Require Recent window $R$, observation window $W$, refresh interval $J$, shared budget $M$
\Ensure Final runtime prediction $\boldsymbol{z}$, or an explicit execution error
\For{each scheduled prefill or decode forward of invocation $a$}
    \State Reconcile $a$'s private cache and reusable prefix; retain sealed block widths
    \If{the resident set changed or a fresh observation is available}
        \State Compute $s_{a,\ell}$ from ordinary-forward $\boldsymbol{Q}/\boldsymbol{K}/\boldsymbol{V}$ for each observed layer
        \State Impute missing layer scores from the resident median, or $1$ if none exists
        \For{each resident invocation $a$}
            \State Sort its $L$ layers by score, with deterministic tie breaking
            \For{$k=1,\ldots,\lfloor(L-1)/2\rfloor$}
                \State Assign $k$ lowest layers $b_{\mathrm{avg}}-1$, $k$ highest $b_{\mathrm{avg}}+1$, and the rest $b_{\mathrm{avg}}$
                \State Evaluate the score surrogate in Eq.~(\ref{eq:aq_estimated_error})
            \EndFor
            \State Retain the minimum-surrogate layer vector; charge its reserved bytes
        \EndFor
        \State Record the selected reservation and average-width byte envelope
    \EndIf
    \State Run the ordinary forward with sealed blocks at their stored widths and the recent raw tail
    \State Append new keys and values at native precision
    \For{each newly eligible complete block $j$ and KV layer $\ell$}
        \State Seal $\mathcal{B}_{a,j}^{(\ell)}$ using current $b_{a,\ell}$ and record its physical width
    \EndFor
    \State Continue agent actions; release completed invocations and their private caches
\EndFor
\State Audit candidate-width coverage, sealed-code mean, and physical KV compression
\State \Return the final prediction $\boldsymbol{z}$
\end{algorithmic}
\end{algorithm}

\noindent
In Algorithm~\ref{alg:aq_online_allocation}, selected widths are prospective: a score refresh changes only subsequent block writes, while sealed blocks remain immutable. Each invocation's balanced layer vector has all three candidate widths and mean $b_{\mathrm{avg}}$. The runtime checks the effective byte envelope at allocation events and records the widths actually written by the packed codec. Accuracy cells are reported only after full-cohort execution and the post-run quantization audit pass.

\section{Details of Baselines, Models and Benchmarks}

\subsection{Details of Baselines}
\label{sec:Details_of_Baselines}

We describe the three quantization baselines in Tables~\ref{tab:main_results} and~\ref{tab:bitwidth_results}, together with the unquantized Vanilla reference.

\begin{itemize}

\item \textbf{Vanilla (FullKV)}.
Vanilla stores the KV cache without quantization and provides the native-precision reference. Table~\ref{tab:bitwidth_results} includes the Qwen3-32B reference with 16-bit KV states, and the controlled memory and latency analysis uses BF16 KV storage under the label FullKV.

\item \textbf{KIVI}~\citep{kivi}.
KIVI is a tuning-free asymmetric KV-cache quantization method. It quantizes keys per channel to isolate the error caused by persistent channel-wise outliers and values per token to reduce attention-output error. KIVI retains a small recent window at the model's native precision and quantizes older entries in groups, supporting streaming updates during autoregressive decoding without model retraining.

\item \textbf{TurboQuant}~\citep{zandieh2026turboquant}.
TurboQuant is a data-oblivious online vector quantizer. It first applies a random rotation that concentrates the coordinate distributions and then uses distribution-aware scalar quantizers. To correct bias in inner-product estimation, TurboQuant stores a one-bit Quantized Johnson--Lindenstrauss (QJL) sketch of the quantization residual. The method requires neither model training nor calibration data and targets both KV-cache storage and attention-score computation.

\item \textbf{PM-KVQ}~\citep{liu2026pm}.
PM-KVQ targets the cumulative error that low-bit quantization introduces during long chain-of-thought decoding. It begins with 16-bit KV states and progressively reduces the precision of the existing cache through 16-to-8-bit, 8-to-4-bit, and 4-to-2-bit transitions as the sequence grows. An equivalent-right-shift operator performs each transition without full dequantization. A calibration stage profiles transformer-block sensitivity, assigns block-specific final bit widths under a memory budget, and uses positional interpolation to approximate long-context key distributions from short sequences.

\end{itemize}

\subsection{Details of Models}
\label{sec:Details_of_Models}

We evaluate three open-weight model backbones covering dense and hybrid-attention architectures: Qwen3-32B, Qwen3.8-27B, and Gemma-4-31B-it.

\begin{itemize}

    \item \textbf{Qwen3-32B}~\citep{yang2025qwen3}.
    Qwen3-32B is a 32B-parameter dense model from the Qwen3 family. Qwen3 unifies thinking and non-thinking modes within a single model and supports adjustable thinking budgets, allowing inference-time computation to be adapted to task difficulty. The family is trained for broad capabilities including mathematical reasoning, coding, instruction following, multilingual understanding, and agent tasks.

    \item \textbf{Qwen3.8-27B}~\citep{qwen38_27b}.
    Qwen3.8-27B is a 27B-parameter native vision-\allowbreak language model with a hybrid-attention backbone. Its 64 language-model layers follow a repeated structure in which three Gated DeltaNet linear-attention layers are followed by one gated full-attention layer, resulting in 48 linear-attention layers and 16 full-attention layers. The model supports a native context length of 262K tokens, multi-token prediction, configurable reasoning effort, and long-horizon agentic execution.

    \item \textbf{Gemma-4-31B-it}~\citep{team2026gemma}.
    Gemma-4-31B-it belongs to the Gemma 4 family of open-weight, natively multimodal language models. Gemma 4 is designed to improve reasoning, long-context capability, and inference efficiency, and introduces an explicit thinking mode that allows the model to generate internal reasoning traces before producing its response. We use the 31B instruction-tuned variant as an additional large-scale backbone from a model family distinct from Qwen.

\end{itemize}

\subsection{Details of Benchmarks}
\label{sec:Details_of_Benchmarks}

We evaluate four benchmarks covering knowledge-\allowbreak intensive search, agentic reasoning, mathematical reasoning, and scientific reasoning.

\paragraph{Knowledge-intensive Search.}

\begin{itemize}

    \item \textbf{Bamboogle}~\citep{press2022compositionality}.
    Bamboogle was introduced to study the compositionality gap of language models. Its questions require multiple inferential steps and are deliberately constructed so that the required facts are unlikely to have appeared together during pretraining. Solving the benchmark therefore requires a model to retrieve or infer individual facts and correctly compose them into the final answer.

\end{itemize}

\paragraph{Agentic Reasoning.}

\begin{itemize}

    \item \textbf{GAIA}~\citep{gaia}.
    GAIA evaluates general AI assistants using real-world questions that require combinations of reasoning, web browsing, multimodal understanding, and tool use. The benchmark contains tasks of varying difficulty and is designed around problems that are straightforward for humans but difficult for AI systems without reliable multi-step interaction with external environments.

\end{itemize}

\paragraph{Mathematical Reasoning.}

\begin{itemize}

    \item \textbf{AIME25}~\citep{aime25}.
    AIME25 consists of problems from the 2025 American Invitational Mathematics Examination. These competition-level problems require multi-step mathematical reasoning across areas such as algebra, geometry, number theory, and combinatorics, with each problem having an exact integer answer.

\end{itemize}

\paragraph{Scientific Reasoning.}

\begin{itemize}

    \item \textbf{GPQA}~\citep{gpqa}.
    GPQA (Graduate-Level Google-Proof Q\&A) contains expert-written multiple-choice questions in biology, physics, and chemistry. The questions are intentionally designed to be difficult to answer through simple web search and instead require graduate-level domain knowledge and multi-step scientific reasoning.

\end{itemize}

\section{Additional Experimental Analysis}

\subsection{Hyperparameter Sensitivity}\label{app:sensitivity_hyperparameters}

We study the observation-window length $W$ and sensitivity-refresh interval $J$ on the AIME25 dataset using Qwen3-32B with an average 4-bit KV-cache budget. Each candidate changes one parameter at a time. Table~\ref{tab:wj_sensitivity} reports the results for all four settings.

\begin{table}[htbp]
\centering
\small
\setlength{\tabcolsep}{5pt}
\renewcommand{\arraystretch}{1.08}
\caption{Hyperparameter sensitivity on AIME25.}
\label{tab:wj_sensitivity}
\begin{tabular}{@{}ccc@{}}
\toprule
$W$ & $J$ & Accuracy (\%) \\
\midrule
16 & 128 & 33.33 \\
\rowcolor{softcyan} \textbf{32} & \textbf{128} & \textbf{40.00} \\
64 & 128 & \textbf{40.00} \\
32 & 64 & \textbf{40.00} \\
\bottomrule
\end{tabular}
\end{table}

We use $(W,J)=(32,128)$ as the default setting because it achieves the highest observed accuracy of 40.00\% while avoiding unnecessary online overhead. Increasing the observation window from $W=32$ to $W=64$ does not improve accuracy, indicating that 32 recent tokens already provide sufficient information for estimating layer sensitivity. Likewise, reducing the refresh interval from $J=128$ to $J=64$ produces the same accuracy but requires sensitivity estimation and allocation updates twice as frequently. In contrast, reducing the observation window to $W=16$ decreases accuracy to 33.33\%, suggesting that an overly short window provides a less reliable estimate of the current attention behavior. We therefore adopt $(W,J)=(32,128)$ throughout the main experiments as it matches the best observed accuracy with a shorter observation window and less frequent sensitivity updates.

\subsection{Online Allocation Ablation}\label{app:online_allocation_ablation}

We isolate the contribution of online sensitivity-based allocation on AIME25 by comparing the default online policy with uniform allocation and fixed sensitivity scores. All runs use Qwen3-32B, an average 4-bit KV-cache budget, and the same byte constraint. Table~\ref{tab:online_allocation_ablation} reports the completed and audited results.

\begin{table}[htbp]
\centering
\small
\setlength{\tabcolsep}{6pt}
\renewcommand{\arraystretch}{1.08}
\caption{Online-allocation ablation on AIME25.}
\label{tab:online_allocation_ablation}
\begin{tabular}{@{}lc@{}}
\toprule
Allocation policy & Accuracy (\%) \\
\midrule
\rowcolor{softcyan} Online sensitivity updates & 40.00 \\
Uniform allocation & 36.67 \\
Fixed sensitivity & 36.67 \\
\bottomrule
\end{tabular}
\end{table}

The online sensitivity-based policy achieves 40.00\% accuracy, outperforming both uniform allocation and fixed sensitivity by 3.34\% under the same average bit budget and byte constraint. Uniform allocation ignores differences in quantization sensitivity across layers, while fixed sensitivity preserves an initial allocation but cannot respond to changes in the agent's queries, histories, and attention behavior during execution. The consistent degradation of these two ablations shows that the benefit of \method{} does not arise solely from using mixed precision; dynamically updating sensitivity estimates and reallocating precision according to the current computation is important for preserving task accuracy. These results support the central design of \method{}, in which online Q/K/V observations continually guide where higher and lower KV-cache precision should be assigned.

\end{document}